\documentclass[11pt]{article}

\usepackage[preprint]{acl}

\usepackage{times}
\usepackage{latexsym}

\usepackage[T1]{fontenc}

\usepackage[utf8]{inputenc}

\usepackage{microtype}

\usepackage{inconsolata}
\usepackage{microtype}
\microtypecontext{spacing=nonfrench}
\usepackage{amssymb}
\usepackage{hyperref}
\usepackage{longtable}
\usepackage{arydshln}
\usepackage{bm}      
\usepackage{booktabs}
\usepackage{array}
\usepackage{makecell}
\usepackage{amsmath, amssymb, amsthm}
\usepackage{cleveref}  
\usepackage{amsmath, etoolbox}
\usepackage{algorithm}
\usepackage{algorithmic}
\usepackage{wrapfig}
\usepackage{makecell}
\usepackage{graphicx}
\usepackage{ragged2e} 
\usepackage{graphicx}
\usepackage{xspace}
\usepackage{multirow}
\usepackage{pifont} 
\usepackage{colortbl} 
\PassOptionsToPackage{table}{xcolor} 
\usepackage{xcolor}
\usepackage{subcaption}
\usepackage{comment}
\usepackage{tabularx}
\usepackage{microtype}      
\usepackage{xcolor}         
\usepackage{hyperref}
\usepackage{longtable}
\usepackage{arydshln}
\usepackage{bm}      
\usepackage{booktabs}
\usepackage{array}
\usepackage{makecell}
\usepackage{amsmath, amssymb, amsthm}
\usepackage{cleveref}  
\usepackage{amsmath, etoolbox}
\usepackage{subcaption}
\usepackage{ragged2e} 
\usepackage{graphicx}
\usepackage{bbm}
\usepackage{xspace}
\usepackage{multirow}
\usepackage{pifont} 
\usepackage{colortbl} 
\usepackage[most]{tcolorbox}
\newcommand{\methodname}{\textsc{Ecpo}}

\title{When Denser Credit Is Not Enough: Evidence-Calibrated Policy Optimization for Long-Horizon LLM Agent Training}

\author{Yuanfan Li\textsuperscript{$1,\dagger$}, Qi Zhou\textsuperscript{$2,\dagger$}, Wenjing Duan\textsuperscript{$2,\dagger$}, Lu Chen\textsuperscript{$1,\ast$} \\
        \textsuperscript{1}
        X-LANCE Lab, School of Computer Science, Shanghai Jiao Tong University, Shanghai, China\\ 
        \textsuperscript{2}Faculty of Electronic and Information Engineering, Xi'an Jiaotong University \\
        \textsuperscript{$\dagger$} Equal contribution, \textsuperscript{$\ast$} Correspondence to: chenlusz@sjtu.edu.cn\\
        }

\begin{document}
\maketitle
\begin{abstract}
Long-horizon LLM agents require reinforcement learning methods that can assign credit to intermediate decisions under sparse and delayed rewards. Recent group-based methods such as GiGPO improve over GRPO by constructing step-level advantages at repeated anchor states. However, we show that such dense credit can be statistically unreliable: under limited rollouts, rare but lucky actions may receive overly large advantages, producing divergent anchor bias and late-stage training oscillation. We propose \underline{E}vidence-\underline{C}alibrated \underline{P}olicy \underline{O}ptimization (\methodname{}), a critic-free policy optimization algorithm that calibrates step-level credit before policy updates. \methodname{} combines Evidence-Calibrated Action Advantage, which groups rollouts by canonical actions and shrinks low-count estimates, with Variance-Gated Credit Weighting, which suppresses anchor states dominated by within-action noise. Experiments on ALFWorld and WebShop with Qwen2.5-1.5B/7B show that \methodname{} consistently outperforms strong baselines, improving GiGPO by \textbf{+5.2}/\textbf{+7.3} success points on ALFWorld/WebShop with Qwen2.5-1.5B while adding only \textbf{0.1\%} additional advantage-computation overhead.
\end{abstract}

\section{Introduction}

Large language models (LLMs) have achieved remarkable success in diverse NLP tasks~\cite{vaswani2017attention, brown2020language, singh2025openai}, and there is increasing interest in deploying them as autonomous agents capable of long-horizon decision-making~\cite{bai2022training,shao2024deepseekmath,feng2025group}. Such \emph{agentic} LLMs interact with complex environments, execute multi-step plans, and must cope with delayed rewards. These capabilities pose significant challenges for reinforcement learning (RL), particularly in assigning credit to intermediate decisions in long-horizon trajectories~\cite{ pignatelli2023survey,wei2025webagent}.

\begin{figure}[t]
    \centering
    \resizebox{0.48\textwidth}{!}{\includegraphics{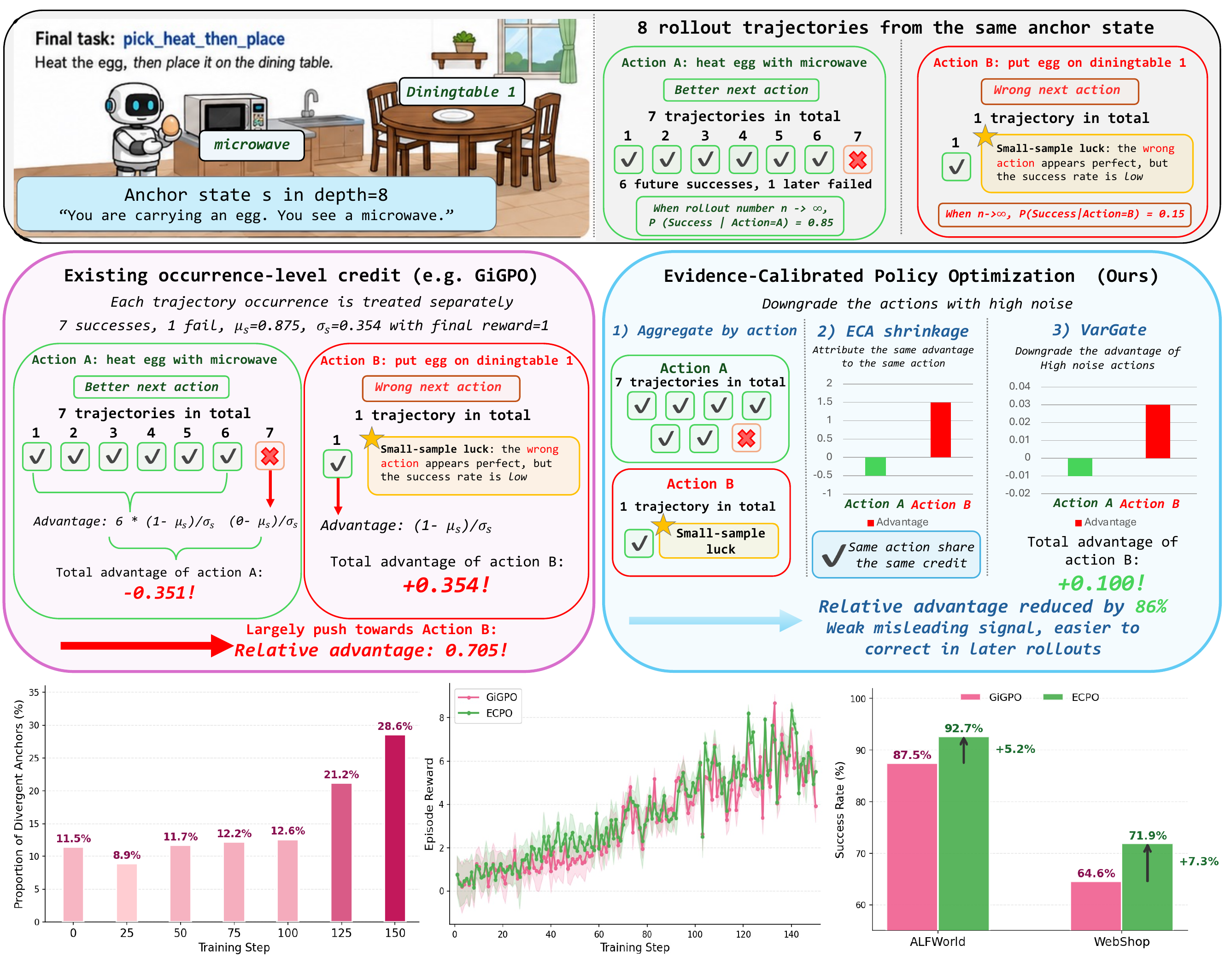}}
    \caption{
    \textbf{Motivation and overview of \methodname.}
    \textbf{Top}: an ALFWorld case illustrating \emph{divergent anchor bias}. We define a divergent anchor as an anchor state where at least one canonical action is sampled more than once, while other actions may have only singleton evidence. Under such imbalanced evidence, GiGPO can over-reward a rare lucky action based on its observed return.
    \textbf{Bottom-left}: divergent anchors become more frequent during training, increasing from 9\% to 28\%.
    \textbf{Bottom-middle}: this bias leads to late-stage reward oscillation; \methodname{} reduces the final reward standard deviation from $\sigma=0.746$ to $\sigma=0.555$.
    \textbf{Bottom-right}: by calibrating action-level evidence, \methodname{} improves final success rates over GiGPO on both ALFWorld and WebShop.
    }
    \label{fig:compare}
    \vspace{-1.2em}
\end{figure}

GRPO~\cite{shao2024deepseekmath} optimizes policies with group-relative trajectory returns, but its credit signal remains tied to final episode outcomes and is therefore sparse in long-horizon agentic tasks. GiGPO~\cite{feng2025group}, an agent-oriented variant of GRPO, mitigates this sparsity by assigning step-level credit at repeated \emph{anchor states}: for rollouts that revisit the same state, it compares their subsequent returns and assigns relative advantages to the actions taken from that state. This enables intermediate decisions to receive credit according to their apparent contribution to eventual task success.

However, Figure~\ref{fig:compare} (bottom middle) shows that denser step-level credit does not automatically translate into stable optimization. Despite providing intermediate supervision beyond sparse terminal rewards, GiGPO exhibits pronounced late-stage reward oscillations, with a final reward standard deviation of $\sigma=0.746$. This observation suggests that the key limitation may lie not only in credit sparsity, but also in the reliability of the constructed step-level credit signal.

To understand the source of this instability, we analyze anchor-level credit assignment in GiGPO and identify a failure mode that we call \emph{divergent anchor bias}. Figure~\ref{fig:compare} (top) shows a representative ALFWorld anchor state with 8 rollouts. At this state, action A is sampled by seven rollouts and leads to six eventual successes, while action B is sampled only once and happens to succeed. Since GiGPO estimates step-level advantages directly from observed downstream returns, the rare action B can receive an overly large positive advantage, even though its true downstream success probability may be lower than that of action A under sufficient sampling. In this case, GiGPO confuses a lucky observation with reliable action quality, thereby pushing the policy toward a high-variance and potentially suboptimal action.

This bias becomes more severe as training progresses. As the policy improves, action distributions at anchor states become increasingly concentrated: dominant actions may appear many times, while low-probability exploratory actions are still occasionally sampled. Such imbalanced anchor groups make rare actions especially vulnerable to overestimation when they happen to succeed. As shown in Figure~\ref{fig:compare} (bottom left), the fraction of divergent anchors increases from 9\% to 28\% during training. This explains the late-stage reward oscillation in Figure~\ref{fig:compare} (bottom middle): once divergent anchors become frequent, GiGPO injects high-variance step-level gradients into policy updates, slowing convergence and destabilizing the later training process.

The underlying mechanism is statistical. Step-level credit assignment aims to identify, at each anchor state, which action has the highest expected downstream success. Yet the true posterior success probability of each action is unknown. With a limited rollout budget, the algorithm can only approximate it using empirical outcomes from a small number of samples. When an action is observed only once, its empirical success rate is either 0 or 1, which is a highly unreliable point estimate. Therefore, directly converting observed returns into step-level advantages can over-reward rare but lucky actions and under-reward frequent actions with more stable evidence.

Motivated by this diagnosis, we propose \textbf{\underline{E}vidence-\underline{C}alibrated \underline{P}olicy \underline{O}ptimization (\methodname)}, which calibrates step-level credit before using it for policy updates. \methodname{} contains two components. \emph{Evidence-Calibrated Action Advantage} (ECA) groups rollouts by canonical actions and applies shrinkage estimation to obtain calibrated action-level advantages, reducing the overestimation of low-count actions. \emph{Variance-Gated Credit Weighting} (VarGate) decomposes anchor-level return variation into between-action signal and within-action noise, and suppresses step-level credit when the observed action differences are statistically unreliable. Together, these components retain the benefit of dense step-level supervision while reducing the variance introduced by small-sample anchor estimates.

We evaluate \methodname{} on two long-horizon agent benchmarks: ALFWorld~\cite{shridhar2021alfworld} and WebShop~\cite{yao2022webshop} with Qwen2.5-1.5B/7B-Instruct~\cite{hui2024qwen2}. Results show that \methodname{} consistently outperforms prompt-based agents, actor-critic RL baselines~\cite{schulman2017proximal}, and prior group-based policy optimization methods such as GRPO~\cite{shao2024deepseekmath} and GiGPO~\cite{feng2025group}. In particular, with Qwen2.5-1.5B, \methodname{} improves over GiGPO by \textbf{+5.2} points on ALFWorld and \textbf{+7.3} success points on WebShop. With Qwen2.5-7B, \methodname{} further improves the overall ALFWorld success rate from 90.8\% to \textbf{91.9\%}, and improves WebShop success rate from 72.4\% to \textbf{74.7\%}. We also analyze different rollout group sizes and find that \methodname{} consistently improves over GiGPO under $N\in\{4,8,10\}$, with gains of \textbf{+4.7}, \textbf{+5.2}, and \textbf{+3.9} points, respectively. These gains come with negligible computational overhead: \methodname{} adds only \textbf{0.1\%} extra advantage-computation overhead over GiGPO, suggesting that evidence-calibrated credit construction provides a practical and scalable way to stabilize long-horizon LLM agent training.

\section{Related Work}

\noindent \textbf{Reinforcement learning for LLM agents.}
RL has been widely used to train language agents for dynamic, interactive, and long-horizon environments. Early work applied DQN~\cite{mnih2015human} to text-based games and language-conditioned decision making~\cite{narasimhan2015language,he2016deep,hausknecht2020interactive}, while recent benchmarks extend this setting to embodied tasks, web navigation, and application-centered tool execution~\cite{shridhar2021alfworld,yao2022webshop,trivedi2024appworld}. Prompt-based agents such as ReAct~\cite{yao2023react} improve interaction by interleaving reasoning and acting, but lack environment-specific policy learning. Recent RL agents explore hierarchical RL, search-guided optimization, entropy-regularized tuning, memory-efficient PPO variants, and end-to-end multi-turn RL~\cite{zhou2024archer,wang2025ragen}. More recent methods address sparse credit: Tree-GRPO uses tree-structured rollouts~\cite{ji2025tree}, GiGPO constructs step-level advantages from repeated anchor states~\cite{feng2025group}, HCAPO refines credit through hindsight reasoning~\cite{tan2026hindsight}, and RAPO augments exploration with retrieved off-policy traces~\cite{zhang2026rapo}. However, these methods densify credit while assuming the estimated step signal is reliable. In contrast, \methodname{} studies credit reliability itself, showing that small rollout groups induce divergent anchor bias and calibrating action-level evidence before updates.

\noindent \textbf{Reinforcement learning for large language models.}
RL is central to improving LLMs. RLHF trains models to follow human preferences~\cite{ziegler2019fine,stiennon2020learning,ouyang2022training,bai2022training}, with PPO~\cite{schulman2017proximal} widely used despite value-model overhead. To reduce this cost, critic-free estimators such as REINFORCE and RLOO have been revisited~\cite{williams1992simple,ahmadian2024back}. For reasoning, GRPO~\cite{shao2024deepseekmath} estimates group-relative advantages from multiple responses, inspiring scalable algorithms including Dr.GRPO, DAPO, CPPO, and GSPO~\cite{liu2025understanding,yu2025dapo,lin2026cppo,zheng2025group}, with applications to mathematical reasoning, search, and tool use~\cite{shao2024deepseekmath,dong2025toolstar,dong2025arpo}. Our work follows this critic-free group-based paradigm, but targets long-horizon agentic tasks where credit must be assigned across environment-dependent actions rather than whole responses. \methodname{} complements existing methods by introducing evidence-calibrated step-level advantages while preserving efficiency.
\vspace{-0.5em}
\section{Preliminaries}
\label{sec:preliminaries}

\noindent \textbf{Problem setup.}
We consider a long-horizon agentic RL setting, where an LLM agent completes a multi-step task specified by $x\sim p(\mathcal{X})$. At step $t=1,\ldots,T$, the agent observes a state $s_t\in\mathcal{S}$, generates a textual action $a_t\in\mathcal{V}^{\leq n}$, receives a reward $r_t\in\mathbb{R}$, and transits to $s_{t+1}$. A full episode forms a trajectory
\begin{equation}
    \tau =
    \{(s_1,a_1,r_1),\ldots,(s_T,a_T,r_T)\}.
\end{equation}
The policy $\pi_\theta(a_t\mid s_t,x)$ defines the action distribution conditioned on the task description and current state. Since rewards in long-horizon tasks are often sparse or delayed, assigning credit to intermediate actions is challenging.

\noindent \textbf{Group-based RL.}
Given a task $x$, group-based RL samples $N$ trajectories $\{\tau_i\}_{i=1}^{N}$ from the old policy $\pi_{\theta_{\mathrm{old}}}$ and computes advantages from within-group return statistics without learning a separate value function. Let $R_i=R(\tau_i)$ be the return of trajectory $\tau_i$. GRPO~\cite{shao2024deepseekmath} estimates the trajectory-level advantage as
\begin{equation}
\label{eq:grpo_adv}
    A^{\mathrm{GRPO}}_i
    =
    \frac{R_i-\mu_R}{\sigma_R+\epsilon},
    \quad
    \mu_R=
    \frac{1}{N}\sum_{j=1}^{N}R_j,
\end{equation}
where $\sigma_R$ is the standard deviation of $\{R_j\}_{j=1}^{N}$ and $\epsilon$ is a small constant. This critic-free formulation is scalable, but it assigns the same trajectory-level advantage to all actions in an episode.

\noindent \textbf{Step-level credit in GiGPO.}
GiGPO~\cite{feng2025group} introduces step-level credit by exploiting repeated \emph{anchor states}. An anchor state is a state $s$ visited by multiple rollout occurrences in the same trajectory group, allowing actions from the same local context to be compared. For trajectory $\tau_i=\{(s_{i,t},a_{i,t},r_{i,t})\}_{t=1}^{T_i}$, the future return from step $t$ is
\begin{equation}
\label{eq:future_return}
    G_{i,t}
    =
    \sum_{\ell=t}^{T_i}
    \gamma^{\ell-t}r_{i,\ell},
\end{equation}
where $\gamma$ is the discount factor. For anchor state $s$, GiGPO collects its occurrence set
\begin{equation}
    \mathcal{I}_s
    =
    \{(i,t)\mid s_{i,t}=s\}.
\end{equation}
It then normalizes future returns within this anchor group:
\begin{equation}
    A^{\mathrm{anchor}}_{i,t}
    =
    \frac{G_{i,t}-\mu_s}{\sigma_s+\epsilon},
    \quad
    (i,t)\in\mathcal{I}_s,
\end{equation}
where $\mu_s$ and $\sigma_s$ are the mean and standard deviation of
$\{G_{j,k}\mid (j,k)\in\mathcal{I}_s\}$.
The final advantage is
\begin{equation}
    \hat{A}^{\mathrm{GiGPO}}_{i,t}
    =
    A^{\mathrm{GRPO}}_i
    +
    \omega A^{\mathrm{anchor}}_{i,t},
\end{equation}
where $\omega$ controls the strength of step-level credit. GiGPO provides denser supervision than GRPO, but its anchor advantage is computed from raw occurrence-level returns and is therefore sensitive to small or imbalanced anchor groups.

\begin{figure*}[t]
    \centering
    \includegraphics[width=0.99\textwidth]{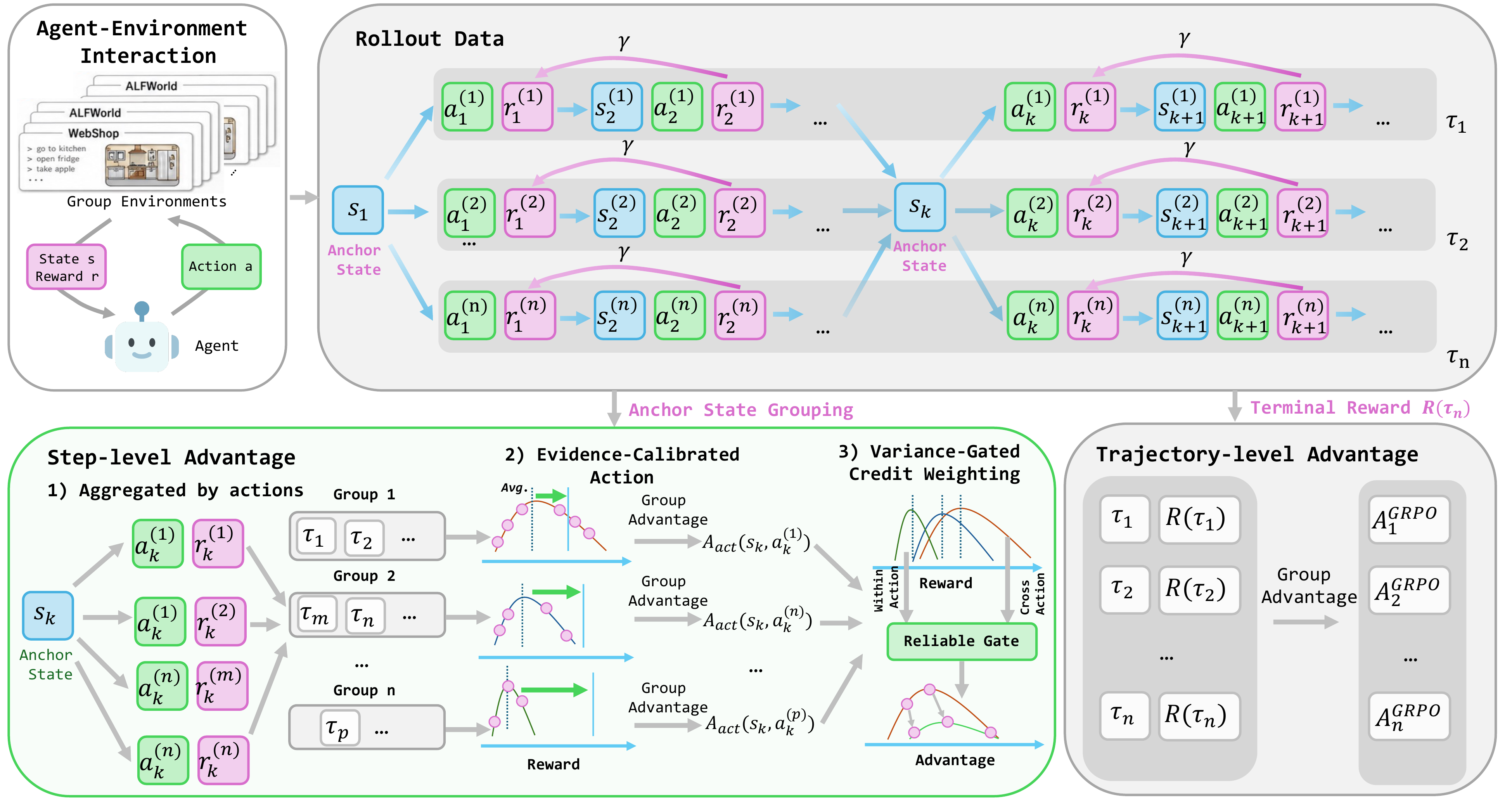}
    \caption{
    \textbf{Overview of \methodname{}.}
    \methodname{} collects grouped rollout trajectories and constructs both trajectory-level and step-level credit.
    At repeated anchor states, rollout occurrences are grouped by canonical actions; \emph{Evidence-Calibrated Action Advantage} shrinks low-count action estimates, while \emph{Variance-Gated Credit Weighting} downweights unreliable anchor signals.
    The calibrated step-level credit is then combined with the trajectory-level group advantage for critic-free policy optimization.
    }
    \label{fig:workflow}
    \vspace{-1em}
\end{figure*}

\section{Methodology}
\label{sec:method}

As discussed in Section~\ref{sec:preliminaries}, GiGPO improves over trajectory-level GRPO by assigning step-level credit at repeated anchor states. However, its anchor advantage $A^{\mathrm{anchor}}_{i,t}$ is computed from raw future returns. Under limited rollouts, this occurrence-level estimator can be unreliable: a rare action with one lucky successful rollout may receive a larger advantage than a frequent action supported by more stable evidence. This produces divergent anchor bias, where the policy update is driven by accidental downstream success rather than reliable action quality, leading to high-variance updates in later training.

We propose \textbf{Evidence-Calibrated Policy Optimization} (\methodname{}), a critic-free algorithm that treats step-level credit construction as a statistical evidence calibration problem. Figure~\ref{fig:workflow} illustrates the overall workflow. Rather than directly trusting occurrence-level returns, \methodname{} aggregates rollouts by canonical actions, calibrates action-level estimates with shrinkage, and estimates anchor reliability through variance decomposition. The final update combines the group-level trajectory advantage with calibrated and reliability-gated step-level credit. In the following subsections, we introduce \emph{Evidence-Calibrated Action Advantage (ECA)}, \emph{Variance-Gated Credit Weighting (VarGate)}, and the resulting policy optimization objective. We analyze \methodname{} theoretically in Appendix~\ref{app:proof}, provide implementation details in Appendix~\ref{app:implementation_details}, and summarize the training procedure in Algorithm~\ref{alg:ecpo}.

\subsection{Evidence-Calibrated Action Advantage}
\label{sec:eca}

GiGPO assigns step-level credit by normalizing the future returns of all occurrences that revisit the same anchor state. This is effective when anchor groups are balanced and sufficiently sampled, but unreliable under limited rollout budgets. When an action is observed only once, its empirical return is a high-variance point estimate: a rare lucky action can receive an overly large advantage even if its true downstream success probability is low. The key issue is that occurrence-level credit confounds current action quality with later trajectory randomness. Thus, step-level credit should be estimated at the action level, and low-count actions should be calibrated before policy updates.

We introduce \emph{Evidence-Calibrated Action Advantage}, which groups occurrences by canonical actions and applies shrinkage estimation. For an anchor state $s$ with occurrence set $\mathcal{I}_s$, each raw textual action $a_{i,t}$ is mapped to a canonical action
\begin{equation}
    u_{i,t}=\operatorname{can}(a_{i,t}),
\end{equation}
where $\operatorname{can}(\cdot)$ is a deterministic environment-grounded canonicalizer that removes superficial textual variation while preserving executable action identity. Invalid or unmatched outputs are mapped to a special invalid-action token rather than merged with valid actions. The canonical action set is
\begin{equation}
    \mathcal{U}_s
    =
    \{u_{i,t}\mid (i,t)\in\mathcal{I}_s\}.
\end{equation}
For each $u\in\mathcal{U}_s$, we define
\begin{equation}
\begin{aligned}
    \mathcal{I}_{s,u}
    &=
    \{(i,t)\in\mathcal{I}_s \mid u_{i,t}=u\}, \\
    n_{s,u}
    &=
    |\mathcal{I}_{s,u}|.
\end{aligned}
\end{equation}

Let $G_{i,t}$ be the future return of occurrence $(i,t)$, and let $\mu_s$ be the mean future return over $\mathcal{I}_s$. The empirical return of action $u$ is
\begin{equation}
    \bar{G}_{s,u}
    =
    \frac{1}{n_{s,u}}
    \sum_{(i,t)\in\mathcal{I}_{s,u}}
    G_{i,t}.
\end{equation}
To avoid over-trusting low-count actions, we shrink the empirical action return toward the anchor-level mean:
\begin{equation}
\label{eq:eca_shrinkage}
    \tilde{\mu}_{s,u}
    =
    \frac{
        n_{s,u}\bar{G}_{s,u}
        +
        \kappa\mu_s
    }{
        n_{s,u}+\kappa
    },
\end{equation}
where $\tilde{\mu}_{s,u}$ is the calibrated action-level return and $\kappa>0$ controls shrinkage strength. Smaller $n_{s,u}$ yields stronger shrinkage toward $\mu_s$, while larger $n_{s,u}$ relies more on $\bar{G}_{s,u}$.

We then normalize calibrated action estimates within the same anchor state. The action-level standard deviation is
\begin{equation}
    \sigma^{\mathrm{act}}_s
    =
    \left(
    \frac{1}{|\mathcal{U}_s|}
    \sum_{u\in\mathcal{U}_s}
    \left(
        \tilde{\mu}_{s,u}-\mu_s
    \right)^2
    \right)^{1/2}.
\end{equation}
The calibrated action advantage is
\begin{equation}
\label{eq:eca_adv}
    A_{\mathrm{act}}(s_{i,t},a_{i,t})
    =
    \frac{
        \tilde{\mu}_{s_{i,t},u_{i,t}}
        -
        \mu_{s_{i,t}}
    }{
        \sigma^{\mathrm{act}}_{s_{i,t}}
        +
        \epsilon
    },
\end{equation}
where $\epsilon>0$ is a numerical constant. All occurrences taking the same canonical action at the same anchor state share the same $A_{\mathrm{act}}$, so step-level credit is based on calibrated action-level evidence rather than noisy individual occurrences.

\subsection{Variance-Gated Credit Weighting}
\label{sec:vargate}

Although ECA reduces low-count overestimation, calibrated action estimates should not always be trusted equally. Some anchor states exhibit clear action-dependent return differences, where the current action strongly explains downstream success. Other anchors are dominated by within-action randomness: even trajectories taking the same action may later diverge and obtain different returns. In such cases, injecting step-level credit can still introduce noisy gradients. Therefore, beyond calibrating action values, \methodname{} must also decide whether an anchor state provides reliable evidence for step-level policy updates.

We introduce \emph{Variance-Gated Credit Weighting}, which decomposes anchor-level return variation into between-action signal and within-action noise. For anchor state $s$, the between-action variance is
\begin{equation}
\label{eq:between_var}
    B_s
    =
    \frac{1}{n_s}
    \sum_{u\in\mathcal{U}_s}
    n_{s,u}
    \left(
        \bar{G}_{s,u}-\mu_s
    \right)^2,
\end{equation}
where $n_s=|\mathcal{I}_s|$. A large $B_s$ indicates that action choice explains substantial return variation.

For each canonical action $u$, the within-action variance is
\begin{equation}
\label{eq:within_var}
    \operatorname{Var}_s(G\mid u)
    =
    \frac{1}{n_{s,u}}
    \sum_{(i,t)\in\mathcal{I}_{s,u}}
    \left(
        G_{i,t}-\bar{G}_{s,u}
    \right)^2.
\end{equation}
The anchor-level within-action variance is
\begin{equation}
    W_s
    =
    \frac{1}{n_s}
    \sum_{u\in\mathcal{U}_s}
    n_{s,u}
    \operatorname{Var}_s(G\mid u).
\end{equation}
Here, $W_s$ captures downstream uncertainty after conditioning on the current action.

To further account for sample size, we define
\begin{equation}
    g_s
    =
    \tanh
    \left(
        \frac{n_s}{\tau}
    \right),
\end{equation}
where $\tau>0$ controls saturation. The final reliability weight is
\begin{equation}
\label{eq:vargate}
    \rho_{\mathrm{VG}}(s)
    =
    g_s
    \cdot
    \frac{
        B_s
    }{
        B_s+W_s+\epsilon
    }.
\end{equation}
When $B_s$ dominates $W_s$, $\rho_{\mathrm{VG}}(s)$ remains large and the step-level signal is trusted; when $W_s$ dominates, the anchor is noisy and its contribution is suppressed. For non-comparable anchors, such as those with fewer than two canonical actions, the reliability weight is set to zero and the update falls back to trajectory-level credit.

\subsection{Policy Optimization with \methodname{}}
\label{sec:ecpo_optimization}

The two components above produce calibrated step-level credit while preserving the stability and scalability of group-based RL. \methodname{} keeps the original trajectory-level group advantage as the global learning signal and injects step-level credit only when the anchor provides reliable action-level evidence. This design retains critic-free optimization, but changes the role of anchor states: they are no longer treated as automatically trustworthy sources of dense supervision, but as statistical evidence whose reliability must be estimated before affecting the policy update.

Let $A^{\mathrm{GRPO}}_i$ denote the group-level trajectory advantage of $\tau_i$. For each occurrence $(i,t)$, \methodname{} constructs the final advantage as
\begin{equation}
\label{eq:ecpo_adv}
\begin{aligned}
\hat{A}^{\methodname}_{i,t}
={}&
A^{\mathrm{GRPO}}_i
+
\omega\,
\rho_{\mathrm{VG}}(s_{i,t})\,
A_{\mathrm{act}}(s_{i,t},a_{i,t}),
\end{aligned}
\end{equation}
where $\omega\geq0$ controls the strength of calibrated step-level credit. When the anchor is unreliable, $\rho_{\mathrm{VG}}(s_{i,t})$ becomes small and the update falls back to trajectory-level group optimization; otherwise, \methodname{} uses the calibrated step signal to refine the update.

The textual action $a_{i,t}$ is generated as a token sequence
$a_{i,t}=(y_{i,t,1},\ldots,y_{i,t,L_{i,t}})$, where $L_{i,t}$ is the action length. Let $c_{i,t,\ell}$ denote the token context of token $y_{i,t,\ell}$, including the task description, interaction history, current state, and previous action tokens. The token-level probability ratio is
\begin{equation}
    q_{i,t,\ell}(\theta)
    =
    \frac{
        \pi_\theta(y_{i,t,\ell}\mid c_{i,t,\ell})
    }{
        \pi_{\theta_{\mathrm{old}}}(y_{i,t,\ell}\mid c_{i,t,\ell})
    }.
\end{equation}

We optimize the policy with the following clipped objective:
\begin{equation}
\label{eq:ecpo_objective}
\begin{aligned}
&\mathcal{J}_{\methodname}(\theta)
=
\mathbb{E}_{i,t,\ell}
\Big[
\min
\Big(
q_{i,t,\ell}(\theta)
\hat{A}^{\methodname}_{i,t},
\\
&\operatorname{clip}
\Big(
q_{i,t,\ell}(\theta),
1-\epsilon_{\mathrm{clip}},
1+\epsilon_{\mathrm{clip}}
\Big)
\hat{A}^{\methodname}_{i,t}
\Big)
\Big]
\\
&-
\beta\,
\mathbb{E}_{i,t,\ell}
\Big[
D_{\mathrm{KL}}
\Big(
\pi_\theta(\cdot\mid c_{i,t,\ell})
\,\Vert\,
\pi_{\mathrm{ref}}(\cdot\mid c_{i,t,\ell})
\Big)
\Big].
\end{aligned}
\end{equation}
where $\epsilon_{\mathrm{clip}}>0$ is the clipping threshold, $\pi_{\mathrm{ref}}$ is the reference policy, $D_{\mathrm{KL}}(\cdot\Vert\cdot)$ denotes the KL divergence, and $\beta\geq0$ is the KL coefficient. The same occurrence-level advantage $\hat{A}^{\methodname}_{i,t}$ is assigned to all tokens of $a_{i,t}$, so the policy promotes or suppresses the entire action according to its evidence-calibrated credit.
\begin{table*}[t]
\centering
\small
\setlength{\tabcolsep}{3.2pt}
\renewcommand{\arraystretch}{1.12}
\caption{
\textbf{Main results on ALFWorld and WebShop.}
For ALFWorld, we report success rates for each subtask as well as the overall result. For WebShop, we report the average task score and success rate. Results are averaged over 3 random seeds. Within each trainable model block, the best result is in \textbf{bold} and the second-best result is \underline{underlined}.
}
\label{tab:main_results}
\resizebox{\textwidth}{!}{
\begin{tabular}{lccccccc|cc}
\toprule
\multirow{2}{*}{\textbf{Method}}
& \multicolumn{7}{c|}{\textbf{ALFWorld}} 
& \multicolumn{2}{c}{\textbf{WebShop}} \\
\cmidrule(lr){2-8} \cmidrule(lr){9-10}
& Pick & Clean & Cool & Look & Heat & Pick2 & All & Score & Success \\
\midrule

\rowcolor{gray!15}
\multicolumn{10}{l}{\textit{Closed-source LLMs}} \\
Gemini-3.5-Flash
& 91.2$_{\pm3.9}$ & 67.1$_{\pm2.5}$ & 33.3$_{\pm6.8}$ & 51.5$_{\pm4.3}$ & 33.3$_{\pm7.8}$ & 57.5$_{\pm1.5}$ & 57.6$_{\pm3.3}$ & 5.8$_{\pm1.4}$ & 3.9$_{\pm0.6}$ \\
GLM-5.1
& 92.9$_{\pm5.2}$ & 82.9$_{\pm1.6}$ & 66.7$_{\pm3.4}$ & 72.7$_{\pm7.4}$ & 85.4$_{\pm5.9}$ & 65.0$_{\pm2.4}$ & 77.3$_{\pm1.7}$ & 12.9$_{\pm4.2}$ & 8.1$_{\pm3.3}$ \\
\midrule

\rowcolor{gray!15}
\multicolumn{10}{l}{\textit{Qwen2.5-1.5B-Instruct}} \\
Prompting
& 5.9 & 3.3 & 4.2 & 5.5 & 9.7 & 0.0 & 4.1 & 23.1 & 5.2 \\
ReAct
& 17.4 & 15.7 & 7.7 & 20.5 & 6.2 & 2.0 & 12.8 & 40.1 & 11.3 \\
Reflexion
& 35.3 & 21.7 & 19.4 & 22.2 & 13.6 & 3.7 & 21.8 & 55.8 & 21.9 \\
PPO
& 64.8$_{\pm3.5}$ & 57.1$_{\pm4.9}$ & 46.4$_{\pm4.0}$ & 40.5$_{\pm6.9}$ & 60.6$_{\pm6.6}$ & 47.4$_{\pm1.9}$ & 54.4$_{\pm3.1}$ & 73.8$_{\pm3.0}$ & 51.5$_{\pm2.9}$ \\
RLOO
& \underline{88.3$_{\pm3.0}$} & 71.0$_{\pm5.9}$ & 66.4$_{\pm5.5}$ & 52.8$_{\pm8.6}$ & 62.8$_{\pm8.7}$ & 56.9$_{\pm4.7}$ & 69.7$_{\pm2.5}$ & 73.9$_{\pm5.6}$ & 52.1$_{\pm6.7}$ \\
GRPO
& 85.3$_{\pm1.5}$ & 84.5$_{\pm6.8}$ & 59.7$_{\pm5.0}$ & 53.7$_{\pm8.0}$ & 78.2$_{\pm7.9}$ & 53.5$_{\pm5.6}$ & 72.8$_{\pm3.6}$ & 64.1$_{\pm5.7}$ & 52.6$_{\pm4.3}$ \\
GiGPO$_{\mathrm{w/std}}$
& 83.2$_{\pm4.6}$ & \underline{94.6$_{\pm4.0}$} & \underline{84.1$_{\pm5.4}$} & \underline{82.2$_{\pm1.6}$} & \underline{84.2$_{\pm0.0}$} & \textbf{91.7$_{\pm4.7}$} & \underline{87.5$_{\pm0.6}$} & \underline{80.4$_{\pm4.2}$} & \underline{64.6$_{\pm3.1}$} \\
\rowcolor{gray!10}
\methodname{}
& \textbf{96.7$_{\pm2.7}$} & \textbf{95.7$_{\pm1.5}$} & \textbf{87.0$_{\pm0.0}$} & \textbf{94.4$_{\pm7.9}$} & \textbf{100.0$_{\pm0.0}$} & \underline{81.7$_{\pm2.4}$} & \textbf{92.7$_{\pm0.7}$} & \textbf{84.4$_{\pm3.2}$} & \textbf{71.9$_{\pm1.7}$} \\
\midrule

\rowcolor{gray!15}
\multicolumn{10}{l}{\textit{Qwen2.5-7B-Instruct}} \\
Prompting
& 33.4 & 19.3 & 2.8 & 21.6 & 6.9 & 3.2 & 14.8 & 26.4 & 7.8 \\
ReAct
& 48.5 & 34.3 & 18.2 & 35.4 & 13.2 & 17.6 & 31.2 & 46.2 & 19.5 \\
Reflexion
& 62.0 & 44.9 & 36.3 & 41.6 & 30.9 & 23.8 & 42.7 & 58.1 & 28.8 \\
PPO
& 92.3$_{\pm4.0}$ & 92.5$_{\pm2.4}$ & \underline{80.3$_{\pm2.0}$} & 64.0$_{\pm8.4}$ & \underline{89.5$_{\pm7.0}$} & 68.8$_{\pm8.3}$ & 80.4$_{\pm2.7}$ & 81.4$_{\pm3.1}$ & 68.7$_{\pm5.1}$ \\
RLOO
& 87.6$_{\pm4.3}$ & 87.3$_{\pm5.8}$ & 71.9$_{\pm5.2}$ & 78.2$_{\pm8.3}$ & 81.3$_{\pm7.6}$ & 48.9$_{\pm8.4}$ & 75.5$_{\pm4.6}$ & 80.3$_{\pm3.2}$ & 65.7$_{\pm4.0}$ \\
GRPO
& 90.8$_{\pm5.1}$ & 89.3$_{\pm5.4}$ & 72.5$_{\pm5.4}$ & 66.1$_{\pm6.7}$ & 74.7$_{\pm6.9}$ & 64.7$_{\pm7.3}$ & 77.6$_{\pm5.2}$ & 79.3$_{\pm2.8}$ & 66.1$_{\pm3.7}$ \\
GiGPO$_{\mathrm{w/std}}$
& \textbf{97.7$_{\pm1.6}$} & \underline{98.8$_{\pm1.6}$} & \textbf{89.3$_{\pm8.2}$} & \textbf{82.7$_{\pm7.9}$} & 83.7$_{\pm7.2}$ & \underline{79.2$_{\pm6.6}$} & \underline{90.8$_{\pm1.3}$} & \underline{84.4$_{\pm0.5}$} & \underline{72.4$_{\pm2.6}$} \\
\rowcolor{gray!10}
\methodname{}
& \underline{94.4$_{\pm1.6}$} & \textbf{100.0$_{\pm0.0}$} & 79.7$_{\pm2.1}$ & \underline{82.2$_{\pm1.6}$} & \textbf{91.2$_{\pm2.5}$} & \textbf{93.3$_{\pm2.4}$} & \textbf{91.9$_{\pm0.7}$} & \textbf{85.5$_{\pm0.5}$} & \textbf{74.7$_{\pm3.0}$} \\
\bottomrule
\end{tabular}
}
\vspace{-1em}
\end{table*}
\section{Experiments and Results}

In this section, we evaluate the effectiveness of \methodname{} on long-horizon agentic tasks and conduct ablation studies to verify the contribution of each component.

\subsection{Results on Long-Horizon Agentic Tasks}
\label{sec:main_results}

\noindent \textbf{Experimental setup.}
We evaluate \methodname{} on two long-horizon agentic benchmarks: ALFWorld~\cite{shridhar2021alfworld} and WebShop~\cite{yao2022webshop}.
For ALFWorld, we report the average success rate including six subtasks: Pick, Clean, Cool, Look, Heat, and Pick2.
For WebShop, we report both the average task score and success rate.
All results are averaged over 3 random seeds.
We use Qwen2.5-1.5B-Instruct and Qwen2.5-7B-Instruct~\cite{hui2024qwen2} as trainable base policies, and compare \methodname{} with three groups of baselines: closed-source LLM agents, including \textit{Gemini-3.5-Flash} and \textit{GLM-5.1}; prompt-based open-source agents, including direct prompting, ReAct~\cite{yao2023react}, and Reflexion~\cite{shinn2023reflexion}; and RL training baselines, including PPO with a critic~\cite{schulman2017proximal}, RLOO~\cite{ahmadian2024back}, GRPO~\cite{shao2024deepseekmath}, and GiGPO~\cite{feng2025group}.
Unless otherwise specified, we follow the training configuration of GiGPO, with a rollout group size of 8.
For \methodname{}, we set the shrinkage strength to $\kappa=2.0$ and the VarGate temperature to $\tau=2.0$.
We further use $d_{\min}=7$ as the depth threshold and $\rho_{\min}=0.5$ as the reliability floor for StrictValid anchors.
Full hyperparameter settings are provided in Appendix~\ref{app:experimental_details}.
\methodname{} is also effective on search-augmented QA tasks, as detailed in Appendix~\ref{sec:qa_results}.
We also analyze the training dynamics in Appendix~\ref{sec:training_dynamics} and the cost of training time in Appendix~\ref{sec:training_time}.

\noindent \textbf{Experiment results.}
Table~\ref{tab:main_results} reports the main results on ALFWorld and WebShop.
We summarize three findings.
\textbf{1) Evidence-calibrated credit improves long-horizon agent performance.}
\methodname{} achieves the best overall performance under the Qwen2.5-1.5B setting.
On ALFWorld, it improves the overall success rate from 87.5 to \textbf{92.7}, and on WebShop it improves the success rate from 64.6 to \textbf{71.9}.
The gains over GiGPO show that denser step-level credit alone is not sufficient; calibrating the reliability of such credit leads to better policy learning in long-horizon environments.
\textbf{2) The gains are consistent with our divergent-anchor diagnosis.}
GiGPO assigns step-level advantages from raw future returns, which can over-reward rare but lucky actions when anchor groups are imbalanced.
\methodname{} directly addresses this issue by shrinking low-count action estimates and gating unreliable anchor states.
This produces more stable optimization: as shown in Figure~\ref{fig:compare}, \methodname{} reduces the final reward standard deviation from $\sigma=0.746$ to $\sigma=0.555$.
Thus, the improvement is not merely due to adding another step-level signal, but to making the step-level signal statistically more reliable.
\textbf{3) \methodname{} remains effective across model scales and tasks.}
The advantage of \methodname{} also holds when scaling to Qwen2.5-7B.
On WebShop, it further improves over GiGPO in both task score and success rate, and on ALFWorld it achieves the best overall 7B result in Table~\ref{tab:main_results}.
These results suggest that evidence-calibrated policy optimization provides a general mechanism for stabilizing group-based RL, especially in settings where repeated anchor states contain imbalanced and noisy action evidence.

\subsection{Ablation Study}
\label{sec:ablation}

\noindent \textbf{Experimental setup.}
We conduct ablation studies on ALFWorld with Qwen2.5-1.5B-Instruct.
All variants follow the same training and evaluation protocol as the main experiments, and results are averaged over 3 random seeds on the \texttt{valid\_seen} split.
We compare four variants: GiGPO as the baseline, \textbf{+ECA} which replaces occurrence-level anchor credit with Evidence-Calibrated Action Advantage but does not use VarGate, \textbf{+VarGate} which applies Variance-Gated Credit Weighting to raw occurrence-level anchor credit without ECA shrinkage, and the full \methodname{} with both components.

\begin{table}[t]
\centering
\scriptsize
\setlength{\tabcolsep}{1.8pt}
\renewcommand{\arraystretch}{1.12}
\caption{
\textbf{Ablation study on ALFWorld with Qwen2.5-1.5B-Instruct.}
We report success rates averaged over 3 random seeds.
``ECA'' and ``VG'' indicate whether Evidence-Calibrated Action Advantage and Variance-Gated Credit Weighting are enabled.
}
\label{tab:ablation}
\resizebox{\columnwidth}{!}{
\begin{tabular}{lccccccccc}
\toprule
\textbf{Method}
& \textbf{ECA}
& \textbf{VG}
& \textbf{All}
& \textbf{Pick}
& \textbf{Pick2}
& \textbf{Look}
& \textbf{Heat}
& \textbf{Cool}
& \textbf{Clean} \\
\midrule
GiGPO
& -- & --
& 87.5$_{\pm0.6}$
& 83.2$_{\pm4.6}$
& \textbf{91.7$_{\pm4.7}$}
& 82.2$_{\pm1.6}$
& 84.2$_{\pm0.0}$
& 84.1$_{\pm5.4}$
& 94.6$_{\pm4.0}$ \\

ECA-only
& \checkmark & --
& 89.8$_{\pm0.6}$
& 88.8$_{\pm1.5}$
& 90.0$_{\pm0.0}$
& 82.2$_{\pm1.6}$
& 93.0$_{\pm2.5}$
& 82.6$_{\pm3.6}$
& 95.6$_{\pm1.6}$ \\

VarGate-only
& -- & \checkmark
& 89.6$_{\pm1.3}$
& 84.3$_{\pm4.0}$
& 90.0$_{\pm4.1}$
& 82.2$_{\pm1.6}$
& 89.5$_{\pm0.0}$
& \textbf{87.0$_{\pm0.0}$}
& \textbf{97.8$_{\pm1.6}$} \\

\methodname{}
& \checkmark & \checkmark
& \textbf{92.7$_{\pm0.7}$}
& \textbf{96.7$_{\pm2.7}$}
& 81.7$_{\pm2.4}$
& \textbf{94.4$_{\pm7.9}$}
& \textbf{100.0$_{\pm0.0}$}
& \textbf{87.0$_{\pm0.0}$}
& 95.7$_{\pm1.5}$ \\
\bottomrule
\end{tabular}
}
\vspace{-2em}
\end{table}

\noindent \textbf{Experiment results.}
Table~\ref{tab:ablation} reports the ablation results.
We summarize three findings.
\textbf{1) Each component independently improves GiGPO.}
Adding ECA alone improves the overall success rate from 87.5 to 89.8, showing that action-level shrinkage effectively reduces the bias of low-count anchor estimates.
Adding VarGate alone improves the overall success rate to 89.6, indicating that variance-based reliability gating can suppress noisy anchor states even without action-level shrinkage.
These results validate the two sources of instability identified in our analysis: small-sample action estimation and unreliable anchor-level credit.
\textbf{2) ECA and VarGate are complementary.}
The full \methodname{} achieves the best overall success rate of \textbf{92.7}, outperforming GiGPO by \textbf{+5.2} points.
This gain is larger than either component alone and also exceeds their simple individual improvements, suggesting a synergistic effect.
ECA first calibrates action-level evidence, while VarGate further decides whether the calibrated step signal is reliable enough to affect the policy update.
\textbf{3) The full method mainly improves tasks sensitive to unstable step credit.}
\methodname{} brings large gains on Pick, Look, and Heat, where long-horizon dependencies make noisy step-level credit especially harmful.
At the same time, Pick2 decreases compared with GiGPO, suggesting that some subtasks may already benefit from the original anchor signal and require less aggressive calibration.
Overall, the ablation confirms our central claim: reliable step-level credit requires both action-level evidence calibration and anchor-level reliability control.
\section{Discussion}

\subsection{Effect of Rollout Group Size}
\label{sec:groupsize_analysis}

\noindent \textbf{Experimental setup.}
We further study how the rollout group size affects step-level credit assignment.
We vary the group size as $N\in\{4,8,10\}$ on ALFWorld with Qwen2.5-1.5B-Instruct, while keeping all other training hyperparameters unchanged.
For each setting, we compare GiGPO and \methodname{} at training step 150 and report the validation success rate.
We also analyze optimization stability using the rolling standard deviation of validation success rate, the maximum single-step success-rate jump, and the average advantage range.

\begin{table}[t]
\centering
\small
\setlength{\tabcolsep}{7pt}
\renewcommand{\arraystretch}{1.12}
\caption{
\textbf{Effect of rollout group size on ALFWorld.}
We evaluate GiGPO and \methodname{} with $N\in\{4,8,10\}$ rollouts using Qwen2.5-1.5B-Instruct.
\methodname{} improves the final success rate across different rollout budgets.
}
\label{tab:groupsize_results}
\begin{tabular}{cccc}
\toprule
\textbf{Group Size $N$} & \textbf{GiGPO} & \textbf{\methodname{}} & \textbf{$\Delta$} \\
\midrule
4  & 71.9\% & \textbf{76.6\%} & +4.7pp \\
8  & 87.5\% & \textbf{92.7\%} & +5.2pp \\
10 & 90.6\% & \textbf{94.5\%} & +3.9pp \\
\bottomrule
\end{tabular}
\vspace{-1em}
\end{table}

\noindent \textbf{Experiment results.}
Table~\ref{tab:groupsize_results} shows that \methodname{} consistently improves over GiGPO under $N\in\{4,8,10\}$, with success-rate gains of \textbf{+4.7}, \textbf{+5.2}, and \textbf{+3.9} points, respectively.
This indicates that evidence-calibrated credit remains effective across different rollout budgets.
Notably, the gain is largest at $N=8$, where GiGPO already has substantially stronger performance than at $N=4$ but still suffers from unstable step-level credit.
This matches our diagnosis: once the policy becomes more capable, repeated anchor states become more frequent, but the sampled evidence at each anchor may still be imbalanced, making rare lucky actions influential in policy updates.
The results also show that simply increasing the number of rollouts does not fully solve the problem.
Although GiGPO improves from $N=4$ to $N=10$, it still remains below \methodname{} at every group size.
Moreover, GiGPO exhibits substantial instability, with the largest single-step jump appearing at $N=8$.
This suggests that intermediate group sizes can produce many anchor signals that are strong enough to affect the policy but not reliable enough to guide it consistently.
\methodname{} mitigates this issue through two complementary mechanisms: ECA suppresses small-sample action bias, especially when rollout evidence is scarce, while VarGate filters noisy anchor states and reduces large policy jumps when more anchor signals appear.
Overall, these results support our central claim that reliable credit construction, rather than simply increasing rollout count, is critical for stable long-horizon agent training.
More details about the impact of rollout group size can be found in Appendix~\ref{app:small_rollout_analysis}.
\section{Conclusion}

We identify divergent anchor bias as a key source of instability in step-level credit assignment for long-horizon LLM agents: denser credit can become harmful when rare lucky actions are over-rewarded under limited rollouts. To address this issue, we propose \methodname{}, which calibrates action-level evidence through shrinkage and gates unreliable anchor states via variance decomposition. Experiments on ALFWorld and WebShop show that \methodname{} improves over GiGPO by \textbf{+5.2} and \textbf{+7.3} success points with Qwen2.5-1.5B, while reducing late-stage reward variance and adding only \textbf{0.1\%} advantage-computation overhead. These results suggest that reliable credit construction is crucial for stable long-horizon agent training.

\section*{Limitations}

Although \methodname{} improves the reliability of step-level credit assignment, it still has several limitations. First, our method relies on repeated anchor states to construct calibrated step-level advantages. In environments where state repetition is rare or where states are difficult to canonicalize, the amount of usable anchor-level evidence may be limited, and \methodname{} may fall back more often to trajectory-level advantages. Second, our experiments focus on ALFWorld, WebShop and SearchQA with Qwen2.5-based agents. While these benchmarks cover embodied planning, web interaction and QA, broader evaluation on more diverse tool-use, multi-agent, and real-world deployment settings is needed to further validate the generality of evidence-calibrated credit assignment.

\section*{Ethics Statement}

This work studies reinforcement learning methods for improving long-horizon LLM agents. The proposed method aims to make policy optimization more stable and reliable by reducing noisy credit assignment, and does not introduce new tools, external data collection, or user-facing autonomous capabilities beyond the evaluated benchmark environments. However, stronger LLM agents may also increase risks when deployed in open-ended settings, especially if they are connected to real tools or services. Therefore, practical deployment should include appropriate safety constraints, human oversight, logging, and environment-specific access control. All experiments are conducted on standard research benchmarks, and no private or sensitive user data is used.

\bibliography{custom}     

\appendix
\section{Experimental Details}
\label{app:experimental_details}

\subsection{Details of Training}
\label{app:training_details}

\noindent \textbf{Hyperparameters for ALFWorld.}
All methods are configured with identical basic hyperparameters for fair comparison.
The maximum prompt length is 2048 tokens, and the maximum response length is 512 tokens.
Each episode allows up to 50 environment steps.
The learning rate is set to $1\times10^{-6}$ for the actor and $1\times10^{-5}$ for the critic, where the critic is used only in PPO.
We adopt a rule-based reward: a reward of 10 is assigned for task success and 0 for failure, and invalid actions receive a penalty of $-0.1$.
For all group-based RL methods, we use a group size of 8 and sample 16 groups per rollout, resulting in $16\times8=128$ parallel environments.
PPO instead uses 128 separate environments for rollouts.
The rollout temperature is set to 1.0, while the validation temperature is set to 0.4.
The mini-batch size is 256, and the KL-divergence loss coefficient is 0.01.
For step-level credit methods, the weighting coefficient $\omega$ is fixed at 1, and the discount factor $\gamma$ is set to 0.95.
For \methodname{}, we set the shrinkage strength to $\kappa=2.0$, the VarGate temperature to $\tau=2.0$, the depth threshold to $d_{\min}=7$, and the reliability floor to $\rho_{\min}=0.5$.

\noindent \textbf{Hyperparameters for WebShop.}
All methods are configured with identical basic hyperparameters.
The maximum prompt length is 4096 tokens, and the maximum response length is 512 tokens.
Each episode is limited to 15 environment steps.
The learning rate is $1\times10^{-6}$ for the actor and $1\times10^{-5}$ for the critic, where the critic is used only in PPO.
We use the same rule-based reward as ALFWorld: 10 for success and 0 for failure, with a penalty of $-0.1$ for invalid actions.
All group-based RL methods use a group size of 8 and sample 16 groups per rollout, totaling $16\times8=128$ parallel environments.
PPO uses 128 distinct environments for rollouts.
The rollout temperature is set to 1.0, while the validation temperature is set to 0.4.
The mini-batch size is 64, and the KL-divergence loss coefficient is 0.01.
For step-level credit methods, the weighting coefficient $\omega$ is fixed at 1, and the discount factor $\gamma$ is set to 0.95.
For \methodname{}, we use the same evidence-calibration hyperparameters as ALFWorld: $\kappa=2.0$, $\tau=2.0$, $d_{\min}=7$, and $\rho_{\min}=0.5$.

\noindent \textbf{Computing details.}
For ALFWorld and WebShop, all Qwen2.5-1.5B experiments are trained on 8 A100-80G GPUs with 12 hours, and all Qwen2.5-7B experiments are trained on 16 A100-80G GPUs with 16 hours.
Each experiment is trained for 150 iterations.
All reported results are averaged over 3 random seeds.

\subsection{Implementation Details of Canonicalization and Anchor Validation}
\label{app:implementation_details}

\noindent \textbf{Action canonicalization.}
The canonicalization function $\operatorname{can}(\cdot)$ is deterministic and environment-grounded.
Given a generated textual action, we first apply lightweight normalization, including lowercasing, trimming whitespace, removing redundant punctuation, and normalizing template variants.
We then match the normalized action against the admissible action set exposed by the environment at the current state.
If the action matches one admissible action, it is mapped to that admissible action string as its canonical form.
If the generated action does not match any admissible action, it is mapped to a special invalid-action token.
Invalid actions are therefore grouped only with other invalid actions and are never merged with valid actions.

\noindent \textbf{Anchor-state matching.}
Anchor states are also constructed deterministically.
When the environment exposes a structured state identifier, we use this identifier as the anchor key.
When only textual observations are available, we use the normalized observation string as the anchor key after applying the same lightweight text normalization used for actions.
Thus, two occurrences are considered to visit the same anchor state only when their canonical state keys match exactly.
This avoids learned or semantic clustering of states and ensures that anchor construction is reproducible.

\noindent \textbf{StrictValid anchors.}
In implementation, calibrated step-level credit is applied only to anchors that pass a lightweight validity check.
An anchor is considered StrictValid if it satisfies four conditions: it appears after the shallow exploration stage, has repeated visits, contains at least two canonical actions, and has non-degenerate return variation.
Formally, for anchor state $s$, we use
\begin{equation}
\begin{aligned}
    \mathrm{StrictValid}(s)
    =
    \mathbf{1}
    \big[
    &\operatorname{depth}(s)\ge d_{\min},
    \,
    n_s\ge2,
    \,
    \\
    & |\mathcal{U}_s|\ge2,
    B_s+W_s>\epsilon
    \big].
\end{aligned}
\end{equation}
If an anchor does not pass this check, we set its step-level reliability weight to zero and the update falls back to the trajectory-level advantage.
For anchors passing the check, we optionally apply a reliability floor:
\begin{equation}
    \rho_{\mathrm{VG}}(s)
    \leftarrow
    \max(\rho_{\mathrm{VG}}(s),\rho_{\min}).
\end{equation}
This prevents reliable anchors from being overly suppressed by small numerical fluctuations.
In our experiments, we set $d_{\min}=7$ and $\rho_{\min}=0.5$.
These two parameters are used only for implementation stability; the core method is defined by the ECA shrinkage estimator and the VarGate reliability weight.

\subsection{Prompt Templates}
\label{app:prompt_templates}

\noindent \textbf{ALFWorld prompt.}
For ALFWorld, we use the following prompt template with recent interaction history and admissible actions.

\begin{tcolorbox}[
    colback=gray!4,
    colframe=gray!45,
    boxrule=0.5pt,
    arc=2mm,
    left=1.5mm,
    right=1.5mm,
    top=1mm,
    bottom=1mm,
    breakable,
    title=\textbf{ALFWorld Prompt Template},
    fonttitle=\small,
    coltitle=black,
    colbacktitle=gray!12
]
\small
You are an expert agent operating in the ALFRED Embodied Environment. Your task is to: \{task\_description\}

\vspace{0.5em}
Prior to this step, you have already taken \{step\_count\} step(s). Below are the most recent \{history\_length\} observations and the corresponding actions you took: \{action\_history\}

\vspace{0.5em}
You are now at step \{current\_step\} and your current observation is: \{current\_observation\}

\vspace{0.5em}
Your admissible actions of the current situation are: [\{admissible\_actions\}].

\vspace{0.5em}
Now it's your turn to take an action.

\vspace{0.5em}
You should first reason step-by-step about the current situation. This reasoning process MUST be enclosed within \texttt{<think>} \texttt{</think>} tags.

\vspace{0.5em}
Once you've finished your reasoning, you should choose an admissible action for current step and present it within \texttt{<action>} \texttt{</action>} tags.
\end{tcolorbox}

\noindent \textbf{WebShop prompt.}
For WebShop, we use the following prompt template with recent interaction history and available actions.

\begin{tcolorbox}[
    colback=gray!4,
    colframe=gray!45,
    boxrule=0.5pt,
    arc=2mm,
    left=1.5mm,
    right=1.5mm,
    top=1mm,
    bottom=1mm,
    breakable,
    title=\textbf{WebShop Prompt Template},
    fonttitle=\small,
    coltitle=black,
    colbacktitle=gray!12
]
\small
You are an expert autonomous agent operating in the WebShop e-commerce environment.

\vspace{0.5em}
Your task is to: \{task\_description\}. Prior to this step, you have already taken \{step\_count\} step(s). Below are the most recent \{history\_length\} observations and the corresponding actions you took: \{action\_history\}. You are now at step \{current\_step\} and your current observation is: \{current\_observation\}. Your admissible actions for the current situation are: [\{available\_actions\}].

\vspace{0.5em}
Now it’s your turn to take one action for the current step. You should first reason step-by-step about the current situation, then think carefully which admissible action best advances the shopping goal. This reasoning process MUST be enclosed within \texttt{<think>} \texttt{</think>} tags.

\vspace{0.5em}
Once you’ve finished your reasoning, you should choose an admissible action for current step and present it within \texttt{<action>} \texttt{</action>} tags.
\end{tcolorbox}

\section{Additional Results}
\subsection{Evaluation on Search-Augmented QA Tasks}
\label{sec:qa_results}

\noindent \textbf{Experimental setup.}
We further evaluate \methodname{} on search-augmented QA tasks, where the agent must decide whether to issue search queries or produce the final answer within a limited number of turns.
Following prior work, we train on NQ~\cite{kwiatkowski2019natural} and HotpotQA~\cite{yang2018hotpotqa}, and evaluate on both in-domain and out-of-domain QA tasks.
The single-hop evaluation includes NQ~\cite{kwiatkowski2019natural}, TriviaQA~\cite{joshi2017triviaqa}, and PopQA~\cite{mallen2023not}, while the multi-hop evaluation includes HotpotQA~\cite{yang2018hotpotqa}, 2WikiMultiHopQA~\cite{ho2020constructing}, MuSiQue, and Bamboogle~\cite{press2023measuring}.
We use Qwen2.5-3B-Instruct as the base model and compare with R1-Instruct~\cite{guo2025deepseek}, Search-R1~\cite{jin2025search}, ZeroSearch~\cite{sun2025zerosearch}, StepSearch~\cite{wang2025stepsearch}, and GiGPO~\cite{feng2025group}.
For fair paired comparison, we report our reproduced GiGPO and \methodname{} results under the same training protocol.

\begin{table*}[t]
\centering
\small
\setlength{\tabcolsep}{4.2pt}
\renewcommand{\arraystretch}{1.12}
\caption{
\textbf{Performance on search-augmented QA tasks.}
Models are trained on NQ and HotpotQA.
$\dagger$ and $*$ indicate in-domain and out-of-domain datasets, respectively.
Best results of each group are bolded.
}
\label{tab:qa_results}
\resizebox{\textwidth}{!}{
\begin{tabular}{lcccccccc}
\toprule
\multirow{2}{*}{\textbf{Method}}
& \multicolumn{3}{c}{\textbf{Single-Hop QA}}
& \multicolumn{4}{c}{\textbf{Multi-Hop QA}}
& \multirow{2}{*}{\textbf{Avg.}} \\
\cmidrule(lr){2-4}
\cmidrule(lr){5-8}
& NQ$^\dagger$ & TriviaQA$^*$ & PopQA$^*$
& HotpotQA$^\dagger$ & 2Wiki$^*$ & MuSiQue$^*$ & Bamboogle$^*$
& \\
\midrule
R1-Instruct
& 27.0 & 53.7 & 19.9
& 23.7 & 29.2 & 7.2 & 29.3
& 27.1 \\
Search-R1
& 34.1 & 54.5 & 37.8
& 32.4 & 31.9 & 10.3 & 26.4
& 32.5 \\
ZeroSearch
& 41.4 & 57.4 & \textbf{44.8}
& 27.4 & 30.0 & 9.8 & 11.1
& 31.7 \\
StepSearch
& -- & -- & --
& 34.5 & 32.0 & \textbf{17.4} & \textbf{34.4}
& -- \\
GiGPO
& \textbf{44.0} & 59.5 & 41.5
& 37.0 & \textbf{37.0} & 12.1 & 26.4
& 42.5 \\
\rowcolor{gray!10}
\methodname{}
& 43.7 & \textbf{60.7} & 44.1
& \textbf{37.6} & 36.6 & 12.7 & 27.2
& \textbf{43.4} \\
\bottomrule
\end{tabular}
}
\end{table*}

\noindent \textbf{Training details.}
The maximum prompt length is 4096 tokens and the maximum response length is 512 tokens.
The maximum interaction turn is 4.
We use a rule-based reward, assigning 1 for a correct final answer and 0 otherwise, with a penalty of $-0.01$ for invalid actions.
The actor learning rate is $1\times10^{-6}$, the training data size is 256, and the rollout group size is 5.
The rollout and validation temperatures are 1.0 and 0.0, respectively.
The mini-batch size is 512, the KL coefficient is 0.001, the step-credit weight is $\omega=1$, and the discount factor is $\gamma=0.95$.
Experiments are trained on 8 A100 GPUs for 200 iterations.

\noindent \textbf{Prompt template.}
We use the following template for search-augmented QA:

\begin{tcolorbox}[
    colback=gray!4,
    colframe=gray!45,
    boxrule=0.5pt,
    arc=2mm,
    left=1.5mm,
    right=1.5mm,
    top=1mm,
    bottom=1mm,
    breakable,
    title=\textbf{Search-Augmented QA Prompt Template},
    fonttitle=\small,
    coltitle=black,
    colbacktitle=gray!12
]
\small
You are an expert agent tasked with answering the given question step-by-step. Your question: \{task\_description\}.

Prior to this step, you have already taken \{step\_count\} step(s). Below is the interaction history where \texttt{<search>} \texttt{</search>} wrapped your past search queries and \texttt{<information>} \texttt{</information>} wrapped the corresponding search results returned by the external search engine. History: \{memory\_context\}

Now it’s your turn to respond for the current step. You should first conduct reasoning process. This process MUST be enclosed within \texttt{<think>} \texttt{</think>} tags. After completing your reasoning, choose only one of the following actions:

(1) If you lack some knowledge, call a search engine using: \texttt{<search>} your query \texttt{</search>}.

(2) If you have enough knowledge to answer confidently, provide your final answer within \texttt{<answer>} \texttt{</answer>} tags, without detailed illustrations. For example, \texttt{<answer>Beijing</answer>}.
\end{tcolorbox}

\noindent \textbf{Experiment results.}
Table~\ref{tab:qa_results} reports the results on search-augmented QA tasks.
Overall, \methodname{} improves over GiGPO from 42.5\% to \textbf{43.4\%}, yielding a \textbf{+1.0} point gain.
The improvement is most visible on out-of-domain single-hop datasets: \methodname{} improves PopQA from 41.5\% to \textbf{44.1\%} and TriviaQA from 59.5\% to \textbf{60.7\%}.
On multi-hop QA, \methodname{} also improves HotpotQA, MuSiQue, and Bamboogle, while being slightly lower on NQ and 2WikiMultiHopQA.

These results suggest that evidence-calibrated credit is useful beyond embodied and web-interaction environments.
Search-augmented QA has shorter horizons than ALFWorld and WebShop, but still requires sequential tool-use decisions: the agent must decide when to search, how to formulate the query, and when to stop with an answer.
\methodname{} provides modest but consistent overall gains, indicating that calibrating noisy step-level evidence remains beneficial when tool-use decisions are sparse and final rewards are delayed.
The smaller gain compared with ALFWorld and WebShop further supports our story: \methodname{} is most beneficial in longer-horizon environments where repeated anchor states, imbalanced action evidence, and noisy step-level credit are more frequent.

\subsection{Training Dynamics Analysis}
\label{sec:training_dynamics}

Final success rates show the endpoint performance of an agent, but they do not reveal whether the policy reaches this performance through stable learning or through noisy late-stage updates.
To examine the optimization process more directly, we analyze both the validation curves and the variance of step-level advantage estimates.
This analysis is designed to support our central claim: the benefit of \methodname{} comes not merely from adding denser step-level credit, but from making such credit statistically more reliable.

\begin{figure*}[t]
    \centering
    \begin{subfigure}[t]{0.5\textwidth}
        \centering
        \includegraphics[width=\linewidth]{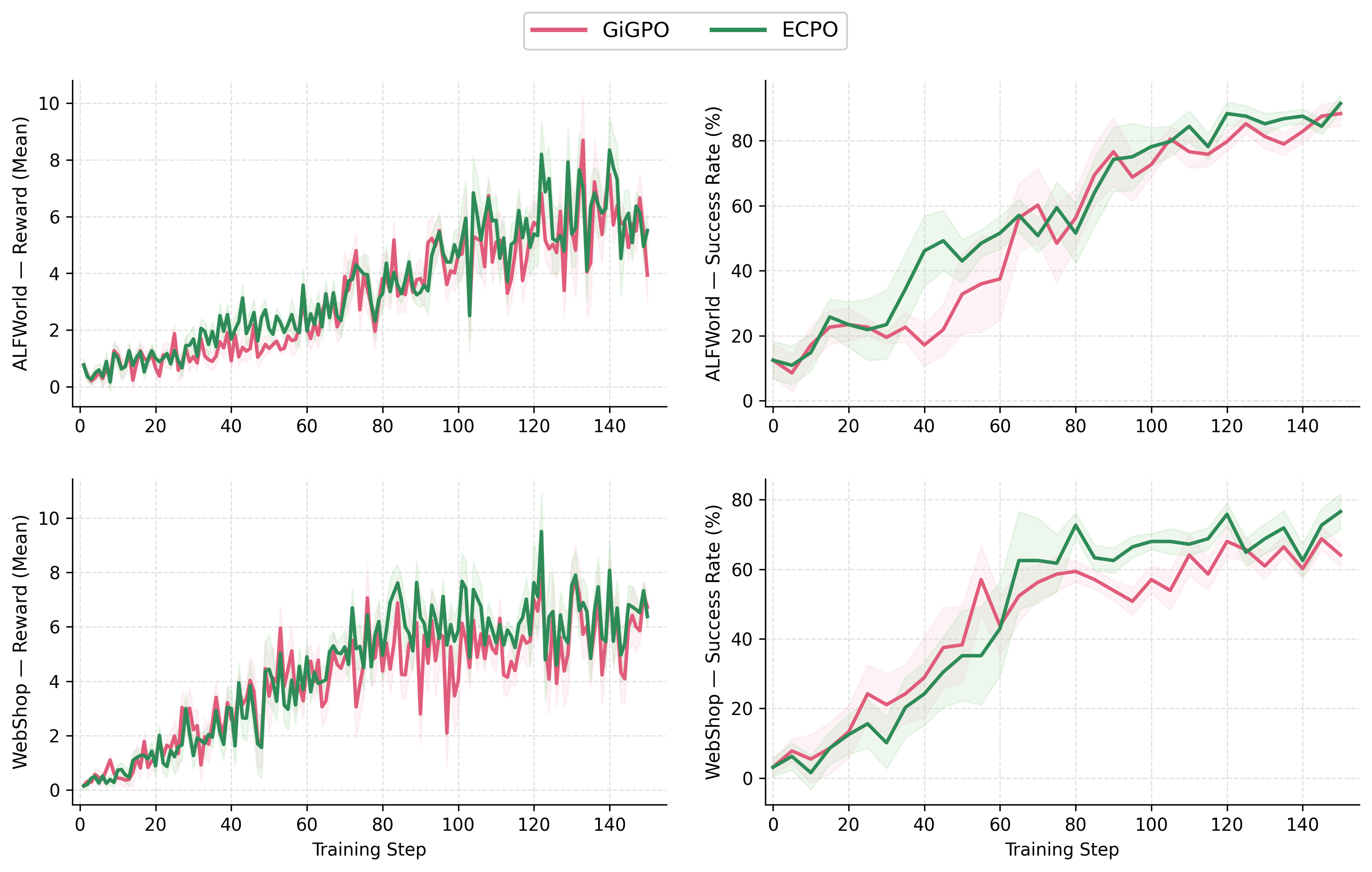}
        \caption{Training dynamics on ALFWorld and WebShop.}
        \label{fig:training_dynamics}
    \end{subfigure}
    \hfill
    \begin{subfigure}[t]{0.49\textwidth}
        \centering
        \includegraphics[width=\linewidth]{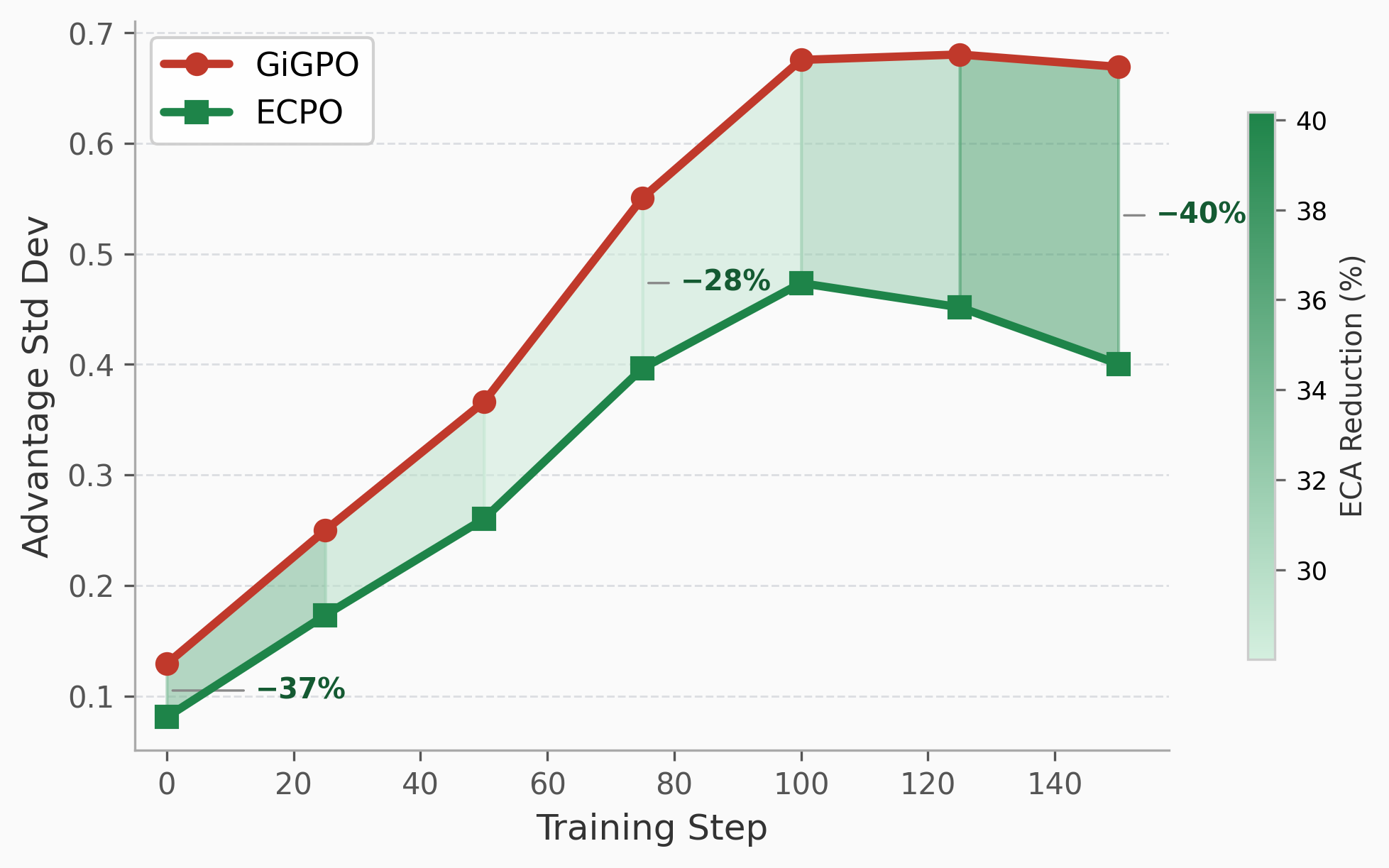}
        \caption{Advantage-variance diagnostic.}
        \label{fig:adv_std_diagnostic}
    \end{subfigure}
    \caption{
    \textbf{Training dynamics and advantage-variance diagnostics.}
    \textbf{Left}: \methodname{} achieves higher final validation performance and more stable reward trajectories on both ALFWorld and WebShop.
    \textbf{Right}: ECA consistently reduces the standard deviation of step-level advantages throughout training, with stronger correction when divergent anchors become more frequent.
    Together, these results show that evidence-calibrated credit improves not only final performance but also the stability of policy optimization.
    }
    \label{fig:training_analysis}
\end{figure*}

\noindent \textbf{Training dynamics.}
Figure~\ref{fig:training_dynamics} compares the learning curves of GiGPO and \methodname{}.
On ALFWorld, the two methods behave similarly in the early stage, but \methodname{} begins to separate from GiGPO after around step 60 and finally reaches 92.7\% validation success rate, compared with 87.5\% for GiGPO.
Its shaded region is also narrower, indicating more stable convergence.
On WebShop, \methodname{} maintains a consistent advantage in the middle and late stages, and the gap continues to enlarge as training proceeds, ending at 71.9\% versus 64.6\%.
This pattern suggests that the benefit of \methodname{} is cumulative: more reliable gradients lead to more persistent policy improvement over training.

The reward curves further support this interpretation.
Compared with GiGPO, \methodname{} achieves a higher reward trajectory with a narrower rolling standard deviation.
This indicates that the policy updates produced by \methodname{} are more directionally consistent and less affected by noisy anchor-level estimates.
Thus, \methodname{} improves not only the final success rate, but also the learning efficiency and stability of the whole training process.

\noindent \textbf{Advantage-variance diagnostic.}
Figure~\ref{fig:adv_std_diagnostic} isolates the effect of ECA on step-level advantage estimation.
The standard deviation of ECA-calibrated advantages remains consistently lower than that of raw GiGPO-style anchor advantages throughout training, showing that ECA is effective across the entire optimization process rather than only at a particular stage.
Although both curves increase as training progresses, reflecting increasingly deterministic policies and larger action-level differences, ECA keeps the advantage variance at a lower level.

More importantly, the reduction becomes stronger again in the later stage.
The relative reduction is 37\% at step 0, decreases to 28\% around step 75, and rises to 40\% by step 150.
This late-stage strengthening coincides with the period where divergent anchors become more frequent, indicating that ECA automatically applies stronger correction when small-sample action estimates become more problematic.
This behavior directly matches the design of ECA: low-count actions receive stronger shrinkage toward the anchor-level mean, preventing rare but lucky actions from producing high-variance advantages.

Overall, Figure~\ref{fig:training_analysis} connects the mechanism and the empirical outcome.
The diagnostic plot verifies that ECA reduces the variance of step-level credit, while the training curves show that this variance reduction translates into smoother optimization and higher final performance.
These results support our story that reliable credit construction, rather than simply denser step-level supervision, is the key to stable long-horizon LLM agent training.

\subsection{Training Time Analysis}
\label{sec:training_time}

\begin{table}[t]
\centering
\small
\setlength{\tabcolsep}{3pt}
\renewcommand{\arraystretch}{1.15}
\caption{
\textbf{Training-time overhead analysis.}
\methodname{} adds only 0.30s for advantage computation over GiGPO, corresponding to about 0.10\% of the total step time. 
Other time differences mainly come from rollout generation and are not caused by the proposed credit-calibration computation.
}
\label{tab:training_time}
\resizebox{\columnwidth}{!}{
\begin{tabular}{lccccc}
\toprule
\textbf{Method}
& \textbf{Adv.}
& \textbf{Gen.}
& \textbf{Actor}
& \textbf{Total}
& \textbf{Extra Cost Source} \\
\midrule
GRPO
& 0.53s
& 183.0s
& 35.2s
& 282.9s
& -- \\
GiGPO
& 0.66s
& 207.7s
& 30.2s
& 302.8s
& -- \\
\methodname{}
& 0.96s
& 232.6s
& 30.0s
& 339.2s
& \textbf{Adv. only: +0.30s} \\
\bottomrule
\end{tabular}
}
\vspace{-0.5em}
\end{table}

\noindent \textbf{Setup.}
We analyze the per-step training time of GRPO, GiGPO, and \methodname{} under the same training infrastructure.
Each training step is decomposed into advantage computation, rollout generation, actor update, and other system overhead.
This analysis isolates whether the evidence-calibrated credit construction in \methodname{} introduces meaningful computational cost.

\noindent \textbf{Training-time analysis.}
Table~\ref{tab:training_time} reports the per-step training time decomposition.
Although \methodname{} introduces additional evidence calibration, its computational overhead is negligible.
Compared with GiGPO, the advantage computation increases only from 0.66s to 0.96s, i.e., an additional 0.30s per step.
This corresponds to only about \textbf{0.10\%} of the total step time, showing that the proposed ECA and VarGate computations add almost no practical overhead.

The increase in total step time should not be interpreted as algorithmic overhead.
Most of the difference comes from the rollout generation stage, which is affected by trajectory length, environment interaction, and sampled response length.
The actor update time remains nearly unchanged, decreasing slightly from 30.2s to 30.0s.
Therefore, the only genuine extra computation introduced by \methodname{} is the lightweight advantage-calibration step, while the overall training pipeline remains essentially as efficient as GiGPO and GRPO.

\subsection{Additional Analysis on Small-Rollout Evidence Calibration}
\label{app:small_rollout_analysis}

To further understand why \methodname{} remains effective under small rollout budgets, we analyze two diagnostic signals: the VarGate reliability weight $\rho_{\mathrm{VG}}$ and the advantage range. These diagnostics complement the group-size discussion in Section~\ref{sec:groupsize_analysis} by showing how \methodname{} adapts its credit correction when the amount of anchor-level evidence changes.

\begin{figure*}[t]
    \centering
    \begin{subfigure}[t]{0.36\textwidth}
        \centering
        \includegraphics[width=\linewidth]{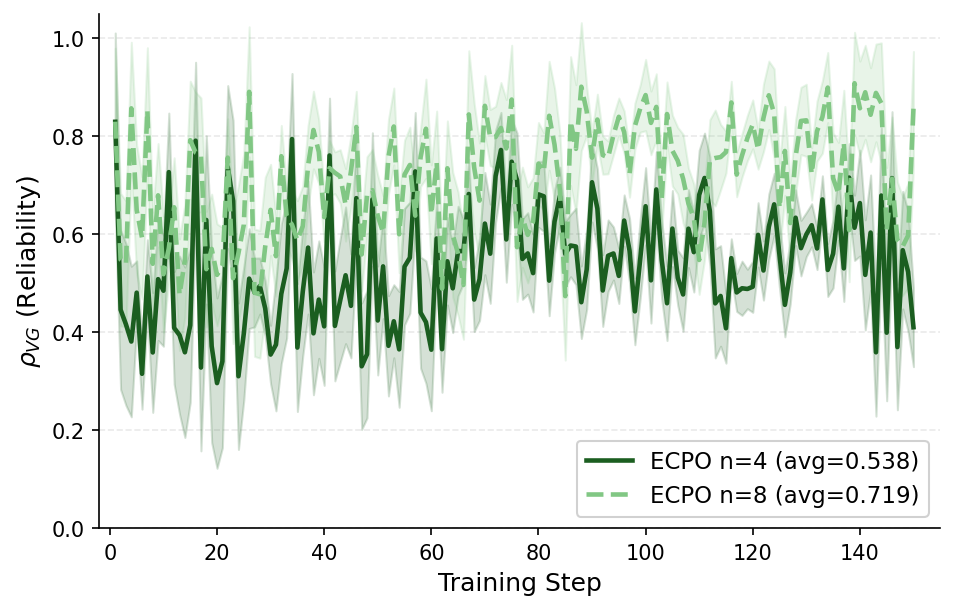}
        \caption{VarGate reliability weight.}
        \label{fig:rho_vs_n}
    \end{subfigure}
    \hfill
    \begin{subfigure}[t]{0.62\textwidth}
        \centering
        \includegraphics[width=\linewidth]{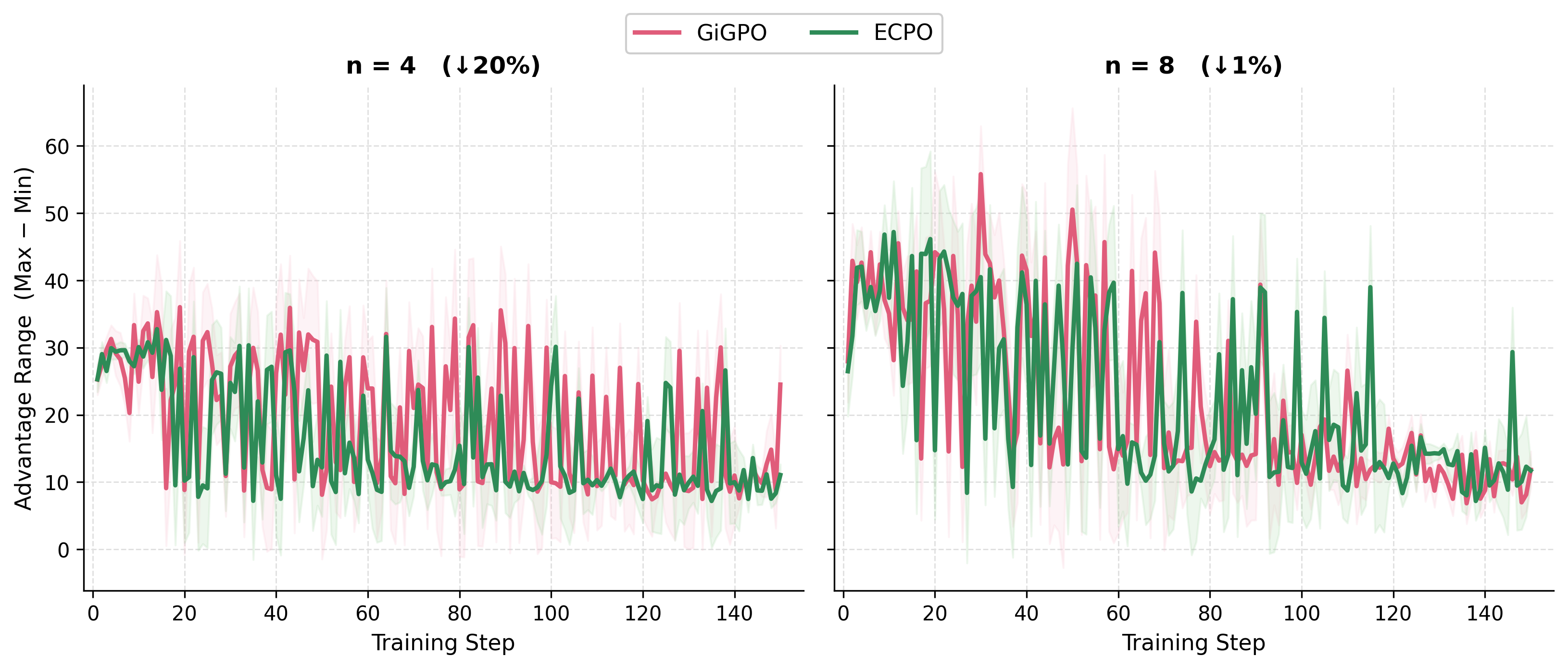}
        \caption{Advantage range.}
        \label{fig:adv_range_vs_n}
    \end{subfigure}
    \caption{
    \textbf{Diagnostics under different rollout group sizes.}
    \textbf{Left}: VarGate assigns lower reliability weights under smaller rollout budgets, indicating that \methodname{} automatically downweights unreliable step-level credit.
    \textbf{Right}: ECA reduces the advantage range more clearly when $N=4$, showing stronger correction under more severe small-sample bias.
    }
    \label{fig:small_rollout_diagnostics}
\end{figure*}

\noindent \textbf{VarGate adapts to rollout evidence.}
Figure~\ref{fig:rho_vs_n} shows the training dynamics of the VarGate reliability weight $\rho_{\mathrm{VG}}$ under $N=4$ and $N=8$.
Since $\rho_{\mathrm{VG}}\in[0,1]$ measures how reliable the anchor-level step signal is, a smaller value means that \methodname{} relies less on step-level credit and more on trajectory-level group advantage.
The $N=4$ curve remains consistently lower than the $N=8$ curve, with average reliability weights of 0.538 and 0.720, respectively.
This indicates that when fewer rollouts are available, each anchor state contains less action-level evidence, and VarGate automatically detects this uncertainty by reducing the contribution of step-level advantages.
Thus, \methodname{} does not require manually changing the step-credit weight for different rollout budgets; its reliability gate provides an adaptive response to finite-sample noise.

\noindent \textbf{ECA is more effective when small-sample bias is stronger.}
Figure~\ref{fig:adv_range_vs_n} compares the advantage range, defined as the maximum advantage minus the minimum advantage within a batch.
A larger range indicates more dispersed advantage estimates, where some actions may be excessively rewarded while others are excessively penalized, leading to less reliable gradient directions.
When $N=4$, GiGPO has an average advantage range of 18.65, while \methodname{} reduces it to 14.97, corresponding to a 20\% reduction.
This shows that ECA shrinkage effectively suppresses extreme estimates caused by low-count actions.
When $N=8$, the ranges of GiGPO and \methodname{} become much closer, 21.35 versus 21.03, suggesting that action-level estimates are already less sparse and leave less room for shrinkage correction.

These diagnostics support the mechanism of \methodname{} under different rollout budgets.
When $N$ is small, the dominant issue is unreliable action estimation; ECA reduces extreme action advantages, while VarGate lowers the overall trust in anchor-level step signals.
When $N$ becomes larger, the action estimates become more stable, so ECA intervenes less aggressively, while VarGate continues to filter noisy anchors.
Therefore, the benefit of \methodname{} is not a fixed heuristic effect: it adapts to the amount and reliability of rollout evidence, providing stronger correction precisely when small-sample bias is more severe.

\section{Pseudo Code of \methodname{}}
\label{app:pseudo_code}

Algorithm~\ref{alg:ecpo} presents the training procedure of \methodname{}.

\begin{algorithm}[t]
    \small
    \caption{\methodname{} Training Procedure}
    \label{alg:ecpo}
    \begin{algorithmic}[1]
        \STATE \textbf{Input:} training task set $\mathcal{D}=\{x\}$,
        policy $\pi_\theta$,
        reference policy $\pi_{\mathrm{ref}}$,
        environment $\mathcal{E}$,
        group size $N$,
        maximum iterations $K$,
        discount factor $\gamma$,
        step-credit weight $\omega$,
        shrinkage strength $\kappa$,
        VarGate temperature $\tau$,
        PPO clipping threshold $\epsilon_{\mathrm{clip}}$,
        KL coefficient $\beta$.
        
        \STATE \textbf{Initialize:} policy parameters $\theta$.
        
        \STATE \parbox[t]{\linewidth}{\centering \textit{*** \methodname{} training begins ***}}
        
        \FOR{$k=1$ to $K$}
            \STATE Set old policy $\pi_{\theta_{\mathrm{old}}}\leftarrow \pi_\theta$.
            
            \FOR{each task description $x\in\mathcal{D}$}
                \STATE \parbox[t]{\linewidth}{\centering \textit{*** Step A: Group rollout collection. ***}}
                
                \STATE Sample $N$ trajectories $\{\tau_i\}_{i=1}^{N}$ from $\pi_{\theta_{\mathrm{old}}}$ in environment $\mathcal{E}$.
                \STATE Compute trajectory returns $\{R_i\}_{i=1}^{N}$.
                \STATE Compute trajectory-level advantages $A^{\mathrm{GRPO}}_i$ by Eq.~\eqref{eq:grpo_adv}.
                
                \STATE \parbox[t]{\linewidth}{\centering \textit{*** Step B: Anchor-state construction. ***}}
                
                \FOR{each trajectory $\tau_i$}
                    \FOR{each step $t=1,\ldots,T_i$}
                        \STATE Compute future return $G_{i,t}$ by Eq.~\eqref{eq:future_return}.
                        \STATE Add occurrence $(i,t)$ into the anchor set $\mathcal{I}_{s_{i,t}}$.
                    \ENDFOR
                \ENDFOR
                
                \STATE \parbox[t]{\linewidth}{\centering \textit{*** Step C: Evidence-Calibrated Action Advantage. ***}}
                
                \FOR{each anchor state $s$}
                    \STATE Canonicalize each action $a_{i,t}$ as $u_{i,t}=\operatorname{can}(a_{i,t})$.
                    \STATE Group occurrences in $\mathcal{I}_s$ by canonical actions $\mathcal{U}_s$.
                    \FOR{each canonical action $u\in\mathcal{U}_s$}
                        \STATE Compute action occurrence set $\mathcal{I}_{s,u}$ and count $n_{s,u}$.
                        \STATE Compute empirical action return $\bar{G}_{s,u}$.
                        \STATE Compute shrinkage-calibrated return $\tilde{\mu}_{s,u}$ by Eq.~\eqref{eq:eca_shrinkage}.
                    \ENDFOR
                    \STATE Compute calibrated action advantage $A_{\mathrm{act}}(s_{i,t},a_{i,t})$ by Eq.~\eqref{eq:eca_adv}.
                \ENDFOR
                
                \STATE \parbox[t]{\linewidth}{\centering \textit{*** Step D: Variance-Gated Credit Weighting. ***}}
                
                \FOR{each anchor state $s$}
                    \STATE Compute between-action variance $B_s$ by Eq.~\eqref{eq:between_var}.
                    \STATE Compute within-action variance $W_s$ by Eq.~\eqref{eq:within_var}.
                    \STATE Compute reliability weight $\rho_{\mathrm{VG}}(s)$ by Eq.~\eqref{eq:vargate}.
                    \IF{$s$ has fewer than two canonical actions}
                        \STATE Set $\rho_{\mathrm{VG}}(s)\leftarrow 0$.
                    \ENDIF
                \ENDFOR
                
                \STATE \parbox[t]{\linewidth}{\centering \textit{*** Step E: Policy optimization. ***}}
                
                \FOR{each occurrence $(i,t)$}
                    \STATE Compute final advantage $\hat{A}^{\methodname}_{i,t}$ by Eq.~\eqref{eq:ecpo_adv}.
                    \STATE Assign $\hat{A}^{\methodname}_{i,t}$ to all tokens of action $a_{i,t}$.
                \ENDFOR
                
                \STATE Optimize $\pi_\theta$ with the clipped objective in Eq.~\eqref{eq:ecpo_objective}.
                
            \ENDFOR
        \ENDFOR
        
        \STATE \textbf{Output:} trained policy $\pi_\theta$.
    \end{algorithmic}
\end{algorithm}

\section{Theoretical Analysis of Evidence Calibration}
\label{app:proof}

This section provides a theoretical analysis of why raw occurrence-level anchor credit can over-reward rare but lucky actions, and why \methodname{} reduces this effect through evidence calibration.

\subsection{Finite Rollouts Cannot Reliably Identify Action Quality}

Consider an anchor state $s$ and a canonical action $u$.
Let the binary downstream success variable after taking action $u$ be
$Y\in\{0,1\}$, with posterior success probability
\begin{equation}
    p_u
    =
    \Pr(Y=1\mid s,u).
\end{equation}
The ideal step-level credit should depend on $p_u$, because $p_u$ measures the true expected downstream success of action $u$ at state $s$.
However, in group-based agentic RL, $p_u$ is unknown and must be estimated from only $n_{s,u}$ rollout occurrences.
When $n_{s,u}$ is small, especially when $n_{s,u}=1$, the empirical estimate
\begin{equation}
    \hat{p}_u
    =
    \frac{1}{n_{s,u}}
    \sum_{j=1}^{n_{s,u}} Y_j
\end{equation}
is a high-variance point estimate.

This limitation is information-theoretic and does not depend on a specific algorithm.
Consider two possible posterior success probabilities $p$ and $p+\Delta$.
From $n$ Bernoulli samples, any method that tries to distinguish
$\mathrm{Bernoulli}(p)$ from $\mathrm{Bernoulli}(p+\Delta)$ has error probability lower bounded by Le Cam's inequality:
\begin{equation}
    P_{\mathrm{err}}
    \geq
    \frac{1}{2}
    \left(
    1-
    \mathrm{TV}
    \big(
    P_p^{n},P_{p+\Delta}^{n}
    \big)
    \right),
\end{equation}
where $P_p^{n}$ denotes the distribution of $n$ i.i.d. Bernoulli samples with success probability $p$.
Using Pinsker's inequality,
\begin{equation}
    \mathrm{TV}
    \big(
    P_p^{n},P_{p+\Delta}^{n}
    \big)
    \leq
    \sqrt{
    \frac{n}{2}
    D_{\mathrm{KL}}
    \big(
    \mathrm{Bern}(p)
    \Vert
    \mathrm{Bern}(p+\Delta)
    \big)
    }.
\end{equation}
For small $\Delta$, the KL divergence satisfies
\begin{equation}
    D_{\mathrm{KL}}
    \big(
    \mathrm{Bern}(p)
    \Vert
    \mathrm{Bern}(p+\Delta)
    \big)
    =
    O
    \left(
    \frac{\Delta^2}{p(1-p)}
    \right).
\end{equation}
Therefore, when
\begin{equation}
    \Delta
    =
    O
    \left(
    \sqrt{
    \frac{p(1-p)}{n}
    }
    \right),
\end{equation}
the two posterior success probabilities cannot be reliably distinguished.
Thus, with small $n_{s,u}$, no distribution-free estimator can consistently determine whether an action is truly good or only lucky from rollout observations alone.
This motivates calibrating the observed action-level evidence before using it for policy optimization.

\subsection{Expected Over-Rewarding of Rare Lucky Actions in GiGPO}

We now analyze a typical divergent anchor.
Assume that at anchor state $s$, there are two canonical actions:
a frequent good action $u^+$ and a rare bad action $u^-$.
The good action is sampled $m$ times, while the bad action is sampled once:
\begin{equation}
    n_{s,u^+}=m,
    \qquad
    n_{s,u^-}=1.
\end{equation}
Let their true posterior success probabilities be
\begin{equation}
\begin{aligned}
    p_+
    &=
    \Pr(Y=1\mid s,u^+), \\
    p_-
    &=
    \Pr(Y=1\mid s,u^-),
    \qquad
    p_+>p_- .
\end{aligned}
\end{equation}
Let $K\sim\mathrm{Binomial}(m,p_+)$ denote the number of successful rollouts among the $m$ occurrences of $u^+$, and let
$Y^-\sim\mathrm{Bernoulli}(p_-)$ denote the observed outcome of the singleton bad action.

GiGPO normalizes raw future returns at the occurrence level.
Conditioned on the event that the bad action succeeds once, i.e., $Y^-=1$, the anchor-level mean return is
\begin{equation}
    \mu_K
    =
    \frac{K+1}{m+1}.
\end{equation}
For binary future returns, the corresponding anchor-level standard deviation is
\begin{equation}
    \sigma_K
    =
    \sqrt{
    \mu_K(1-\mu_K)
    }.
\end{equation}
The GiGPO advantage assigned to the singleton bad action is then
\begin{equation}
    A^{-}_{\mathrm{GiGPO}}(K)
    =
    \frac{1-\mu_K}{\sigma_K+\epsilon}.
\end{equation}
The average GiGPO advantage of the frequent good action is
\begin{equation}
    \bar{A}^{+}_{\mathrm{GiGPO}}(K)
    =
    \frac{K/m-\mu_K}{\sigma_K+\epsilon}.
\end{equation}
Therefore, when the singleton bad action succeeds and the frequent good action is not perfectly successful, i.e., $Y^-=1$ and $K<m$, GiGPO assigns the bad action a higher average step advantage:
\begin{equation}
    A^{-}_{\mathrm{GiGPO}}(K)
    -
    \bar{A}^{+}_{\mathrm{GiGPO}}(K)
    =
    \frac{1-K/m}{\sigma_K+\epsilon}
    >
    0.
\end{equation}
Thus, although $p_-<p_+$, the bad action can receive a larger step-level advantage due to a lucky singleton outcome.

The probability of this over-rewarding event is
\begin{equation}
\label{eq:gigpo_bad_event_prob}
    \Pr
    \left(
    A^{-}_{\mathrm{GiGPO}}
    >
    \bar{A}^{+}_{\mathrm{GiGPO}}
    \right)
    =
    p_-
    \left(
    1-p_+^m
    \right).
\end{equation}
The expected positive advantage gap assigned to the bad action is
\begin{equation}
\label{eq:gigpo_bad_gap}
\begin{aligned}
    \mathbb{E}
    \left[
    \Delta_{\mathrm{GiGPO}}^{-}
    \right]
    =
    p_-
    \sum_{k=0}^{m-1}
    &
    {m\choose k}
    p_+^k
    (1-p_+)^{m-k}
    \\
    &\cdot
    \frac{
    1-k/m
    }{
    \sqrt{\mu_k(1-\mu_k)}+\epsilon
    },
\end{aligned}
\end{equation}
where
\begin{equation}
    \mu_k=\frac{k+1}{m+1}.
\end{equation}
Equation~\eqref{eq:gigpo_bad_gap} shows that the expected erroneous advantage is strictly positive whenever $p_->0$ and $p_+<1$.
Therefore, raw occurrence-level credit can systematically over-reward rare but lucky actions under finite rollout budgets.

\subsection{Expected Reduction under \methodname{}}

\methodname{} reduces this over-rewarding effect in two steps.
First, Evidence-Calibrated Action Advantage shrinks low-count action estimates toward the anchor-level mean.
For the singleton bad action with observed return $Y^-=1$, the ECA-calibrated return is
\begin{equation}
    \tilde{\mu}_{s,u^-}
    =
    \frac{
    1+\kappa\mu_K
    }{
    1+\kappa
    }.
\end{equation}
Hence its centered evidence becomes
\begin{equation}
\label{eq:bad_shrinkage}
    \tilde{\mu}_{s,u^-}-\mu_K
    =
    \frac{
    1-\mu_K
    }{
    1+\kappa
    }.
\end{equation}
Compared with the raw centered evidence $1-\mu_K$ used by GiGPO, the singleton bad action is shrunk by a factor of
\begin{equation}
    \frac{1}{1+\kappa}.
\end{equation}
With the default $\kappa=2$, this factor is $1/3$.

Second, Variance-Gated Credit Weighting multiplies the calibrated action advantage by the reliability weight
\begin{equation}
    \rho_{\mathrm{VG}}(s)
    =
    g_s
    \cdot
    \frac{B_s}{B_s+W_s+\epsilon},
    \qquad
    0\leq \rho_{\mathrm{VG}}(s)\leq 1.
\end{equation}
Thus, under the same local normalization scale, the erroneous singleton bad-action advantage satisfies
\begin{equation}
\label{eq:ecpo_bad_bound}
\begin{aligned}
    \mathbb{E}
    \left[
    \Delta_{\methodname}^{-}
    \right]
    &\lesssim
    \mathbb{E}
    \left[
    \frac{\rho_{\mathrm{VG}}(s)}
    {1+\kappa}
    \Delta_{\mathrm{GiGPO}}^{-}
    \right]
    \\
    &\leq
    \frac{1}{1+\kappa}
    \mathbb{E}
    \left[
    \Delta_{\mathrm{GiGPO}}^{-}
    \right].
\end{aligned}
\end{equation}
Therefore, \methodname{} provably reduces the expected positive advantage assigned to rare lucky actions.
With $\kappa=2$, ECA alone reduces the centered singleton evidence by at least a factor of $3$, and VarGate further suppresses anchors whose observed differences are dominated by within-action noise.

\subsection{Training-Stage Noise Estimated from Validation Success Rate}

The previous analysis shows that the harmful event is driven by singleton lucky successes.
In practice, the probability that a rare action obtains a successful downstream rollout increases as the agent becomes more capable, because later trajectories are more likely to recover from imperfect intermediate actions.
We therefore use the GiGPO validation success rate at training step $t$, denoted by $\mathrm{SR}_t$, as a proxy for the lucky-success noise mass:
\begin{equation}
    p^{-}_t
    \propto
    \mathrm{SR}_t.
\end{equation}
This does not claim that every action has success probability $\mathrm{SR}_t$; rather, it estimates how likely a singleton rollout is to appear successful due to downstream recovery or stochastic continuation.

Table~\ref{tab:noise_proxy} reports five representative training steps.
The singleton lucky-success proxy increases substantially during training, from 0.1250 at step 0 to 0.8828 at step 150.
According to Eq.~\eqref{eq:bad_shrinkage}, ECA reduces this singleton-centered evidence by a factor of $1/(1+\kappa)=1/3$ before VarGate further gates unreliable anchors.

\begin{table}[t]
\centering
\small
\setlength{\tabcolsep}{5pt}
\renewcommand{\arraystretch}{1.12}
\caption{
\textbf{Noise proxy estimated from GiGPO validation success rate.}
We use $\mathrm{SR}_t$ as a proxy for the probability that a singleton rollout appears successful.
With $\kappa=2$, ECA reduces the centered singleton evidence by a factor of $1/3$ before applying VarGate.
}
\label{tab:noise_proxy}
\begin{tabular}{ccc}
\toprule
\textbf{Step} & \textbf{$\mathrm{SR}_t$} & \textbf{ECA-scaled proxy} \\
\midrule
0   & 0.1250 & 0.0417 \\
50  & 0.3281 & 0.1094 \\
75  & 0.4844 & 0.1615 \\
125 & 0.8516 & 0.2839 \\
150 & 0.8828 & 0.2943 \\
\bottomrule
\end{tabular}
\end{table}

This explains why divergent anchor bias becomes more harmful in the later stage of training.
As $\mathrm{SR}_t$ increases, a rare action is more likely to obtain a lucky successful rollout, and GiGPO is more likely to convert this singleton observation into a large positive advantage.
\methodname{} counters this effect by applying stronger shrinkage to low-count actions and suppressing unreliable anchor states through VarGate.
Consequently, even when late-stage training produces more singleton lucky successes, their expected contribution to the policy gradient is substantially reduced.

\end{document}